\documentclass{article}

\usepackage{microtype}
\usepackage{graphicx}
\usepackage{subcaption}
\usepackage{booktabs} % for professional tables
\usepackage{enumitem}
\usepackage{soul}
\usepackage{hyperref}
\usepackage{multirow}

\usepackage[accepted]{icml2023}

\usepackage{amsfonts, amsmath, amssymb, amsthm}
\usepackage{bm}

\newcommand{\mcl}{\mathcal}

\newcommand{\mbf}{\mathbf}
\newcommand{\mbb}{\mathbb}

\newcommand{\bt}{\mbf t}
\newcommand{\bs}{\mbf s}
\newcommand{\bx}{\mbf x}
\newcommand{\bu}{\mbf u}

\newcommand{\bz}{\mbf z}

\newcommand{\bo}{\mbf o}

\newcommand{\eps}{\epsilon}

\newcommand{\bphi}{{\bm{\phi}}}
\newcommand{\bpsi}{{\bm{\psi}}}

\newcommand{\mH}{\mcl{H}}
\newcommand{\mK}{\mcl{K}}
\newcommand{\mF}{\mcl{F}}

\newcommand{\mB}{\mcl{B}}

\newcommand{\mU}{\mcl{U}}
\newcommand{\mV}{\mcl{V}}
\newcommand{\mP}{\mcl{P}}

\newcommand{\R}{\mbb{R}}

\newcommand{\bh}[1]{ {\color{blue} #1} }

\theoremstyle{plain}
\newtheorem{theorem}{Theorem}[section]

\theoremstyle{definition}

\theoremstyle{remark}

\usepackage{todonotes}

\usepackage{amsopn}

\DeclareMathOperator*{\argmin}{arg\,min}

\DeclareMathOperator*{\minimize}{{\rm minimize}}

\icmltitlerunning{A Kernel Approach for PDE Discovery and Operator Learning}

\begin{document}



\twocolumn[
\icmltitle{A Kernel {Framework} for PDE Discovery and Operator Learning}



\icmlsetsymbol{equal}{*}

\begin{icmlauthorlist}
\icmlauthor{Da Long}{utah}
\icmlauthor{Nicole Mrvaljevi{\'c}}{uw}
\icmlauthor{Shandian Zhe}{utah}
\icmlauthor{Bamdad Hosseini}{uw}
\end{icmlauthorlist}

\icmlaffiliation{utah}{School of Computing, University of Utah, UT, USA}
\icmlaffiliation{uw}{Department of Applied Mathematics, University of Washington, WA, USA}

\icmlcorrespondingauthor{Bamdad Hosseini}{bamdadh@uw.edu}
%\icmlcorrespondingauthor{Firstname2 Lastname2}{first2.last2@www.uk}

\icmlkeywords{Kernel methods, Gaussian Processes, Equation discovery, Operator 
Learning}

\vskip 0.3in
]

\printAffiliationsAndNotice{}  % leave blank if no need to mention equal contribution

% \twocolumn[

% %\aistatstitle{Discovering and Solving Partial Differential Equations with Kernels}
% \aistatstitle{A Kernel Approach for PDE Discovery and Operator Learning}

% \aistatsauthor{ Da Long \And Nicole Mrvaljevi{\'c}}

% \aistatsaddress{ School of Computing, \\ University of Utah, UT 
% \And Department of Applied Mathematics, \\ University of Washington, WA} 

%  \aistatsauthor{Shandian Zhe \And Bamdad Hosseini}

% \aistatsaddress{School of Computing, \\ University of Utah, UT \And 
%  Department of Applied Mathematics, \\  University of Washington, WA} 

% ]

\begin{abstract}
  This article presents a three-step framework for learning and solving  partial differential equations (PDEs)  using kernel methods. Given a training set consisting 
  of pairs of noisy PDE solutions and source/boundary terms on a mesh, kernel smoothing is utilized to 
  denoise the data and approximate derivatives of the solution. This information is then used in 
  a kernel regression model to learn the algebraic form of the PDE. The learned PDE is 
  then used within a kernel based solver to approximate the solution of the PDE 
 with a new source/boundary term, thereby constituting an operator learning framework.   Numerical experiments compare the method to state-of-the-art algorithms
 and demonstrate its 
 competitive performance.
\end{abstract}

\section{INTRODUCTION}\label{sec:introduction}
Partial differential equations (PDEs) are 
ubiquitous in natural sciences such as physics \cite{riley1999mathematical}, social sciences \cite{black1973pricing}, biology \cite{edelstein2005mathematical} and engineering \cite{marsden1994mathematical, temam2001navier}. 
To some extent, PDEs are the main subject of interest 
in the field of Physics Informed Learning (PIL); see \cite{karniadakis2021physics, carleo2019machine, willard2020integrating}. 
Traditionally, PDEs are designed or discovered by experts based on mathematical and physical intuition, a process that relies on human expertise, data, and 
mathematical analysis. Once the PDE is accepted as a model it is often solved 
using computer algorithms to simulate a real-world process of interest.

Recent advances in machine learning (ML) along with the abundance of data 
have led to the idea of automating this workflow, thereby promising 
computer programs  for discovering a PDE from limited and noisy data  and 
solving the discovered  
PDE to predict the state of a physical system under previously unseen 
conditions. The goal of this paper is to present an example of such
a workflow based on recent advances {in the theory of kernel methods that is 
robust to noise and supported by theoretical analysis.}

To be precise, we consider the setting where the solution of a PDE subject to known 
forcing is observed at a finite set of locations. This solution is further corrupted by 
noise and constitutes the training data. 
Then we consider two problems: 
(a) {\it Discover the PDE}, i.e.,  
 find the functional relationship between the partial derivatives of 
 the solution that describes the PDE; 
(b) {\it Solve the PDE} subject to 
 previously unseen forcing. 
 Problem (a) is often referred to as   
 {\it equation discovery} and goes back, at least, to the seminal works
 \cite{bongard2007automated, schmidt2009distilling}. More recently, it is often 
 tackled by the Sparse Identification of 
 nonlinear Dynamical Systems (SINDy) algorithm of \cite{brunton2016discovering} 
 and its subvariants. Problem (b) is classical in the field of applied mathematics and 
 numerical PDEs. However, classical numerical methods are very intrusive 
 and are often tailored to specific types of PDEs. 
 Recent techniques such as the Physics Informed Neural Networks (PINNs) of 
 \cite{raissi2019physics} or the kernel approach of \cite{chen2021solving} 
 can circumvent this issue by relaxing the PDE from an equality constraint 
 to a regression misfit term. {Solving Problems (a) and (b) in conjunction yields an
 {\it operator learning} framework}, where one aims to learn/approximate the infinite-dimensional solution operator of a PDE; 
the DeepONet algorithm of \cite{lu2021learning} and the Fourier Neural Operator (FNO) 
approach of \cite{li2020fourier} are state of the art in this context.

{
The main contributions of the article are two-fold:
}
\begin{enumerate}[label=(\arabic*)]
    
    \item {We present a three-step kernel method for the discovery of PDEs and learning of 
    their solution operators from noisy and limited data: Step (i), kernel smoothing 
is utilized to denoise the training data and compute pertinent 
partial derivatives of the solution; Step (ii), kernel regression is 
used to learn the algebraic form of the PDE; Step (iii), the kernel 
solver of \cite{chen2021solving} is used to numerically approximate the solution to the discovered PDE
under new forcing data.}

{Step (i) allows us to 
accommodate input data that are provided on unstructured and inconsistent grids. Step (ii) 
allows us to learn the functional form of PDEs with variable (spatially or temporally) parameters.
Step (iii) is, at the level of implementation, blind to the functional form of the PDE. 
All three steps inherit the desirable robustness and stability properties of kernel methods 
and are amenable to kernel learning strategies such as cross validation or kernel flow 
\cite{owhadi2019kernel}.
}

\item {Our three-step approach is compatible with, and complements, existing 
PDE discovery algorithms and PDE solvers. For example, the
kernel smoothing approach of Step (i) can be used for denoising and gradient estimation 
within the SINDy and PDE-FIND  algorithms of \cite{rudy2017data}, thereby extending these 
methods to  training data  given on unstructured grids. 
In fact, we demonstrate that SINDy can be used in place of our kernel PDE discovery 
approach in step (ii). At the same time 
other meshless PDE solvers such as PINNs can replace the one we use in step (iii).}

% \item \bh{We present theoretical results that demonstrate the asymptotic convergence properties 
% of our three step kernel framework under idealized assumptions. (TBA) }

\end{enumerate}

% Our proposed approach can be summarized in three steps: (1) kernel smoothing 
% is utilized to denoise the training data and compute pertinent 
% partial derivatives of the solution; (2) kernel regression is 
% used to learn the algebraic form of the PDE; (3) the kernel 
% solver of \cite{chen2021solving} is used to solve the discovered PDE under new conditions 
% such as a different forcing than those observed in the training data. 
% Step (1)  is classical in the field of Gaussian process (GP) regression
% and can be used by itself as a denoiser for other methods such as SINDy. Our approach is 
% simple   to implement and allows for learning of the kernel that will be used in step 3. Step (2) can be viewed as a kernel analogue of SINDy although we 
% do not impose any sparsity assumptions on our features. Using the kernel perspective here 
% allows us to have more flexibility and accommodate more features as well as variable coefficient
%  PDEs.
% Step (3) is a meshless PDE 
% solver that can handle  general nonlinearities and allows us to incorporate the PDE 
% as a soft constraint and still be able to solve the learned equation without having to 
% worry about well-posedness of the learned PDE.
% Combining steps 1--3 
% yields an operator learning framework which, compared to state-of-the-art methods, 
% requires significantly less training data by leveraging the knowledge that the 
% operator of interest is the solution map of a PDE.

The rest of the article is organized as follows: Section~\ref{sec:preliminaries}
reviews background material; Section~\ref{sec:methodology}
outlines our proposed methodology in detail as well as some supporting theory;  Section~\ref{sec:experiments}
presents our numerical experiments; Section~\ref{sec:discussion} presents 
a more detailed discussion of our findings and their implementation as well 
as how our methodology relates to existing methods in the literature;
and Section~\ref{sec:conclusion} summarizes 
our conclusions. 
A detailed review of the relevant literature is presented in 
Section~\ref{sec:related-lit} followed by implementation details and additional  experimental results   in Section~\ref{sec:experiment-details}.

\section{PRELIMINARIES}\label{sec:preliminaries}
We collect here some preliminary results and notation that will be used in the remainder of the article.

\subsection{Nonlinear PDEs}
{Suppose $D \ge 1$ and  let $\Omega \subset \R^D$  be a compact and simply-connected domain with boundary
$\partial \Omega$. Consider 
the multi-index  $\bm{\alpha} = (\alpha_1, \dots, \alpha_D) \in \mathbb{N}^D$
(i.e., a $d$-dimensional vector of non-negative integers)\footnote{Henceforth we use 
bold letters to denote  $d$-dimensional vectors of integers or reals for $d \ge 2$.}.
Now for a smooth function $u:\Omega \to \R$ 
we define the partial derivatives 
$\partial^{\bm{\alpha}} u := \partial_{x_1}^{\alpha_1} \partial_{x_2}^{\alpha_2}
\dots \partial_{x_D}^{\alpha_D} u$.
 We further consider two collections of multi-indices
 $ \{ \bm{\alpha}^1, \dots, \bm{\alpha}^{P} \} \subset \mathbb{N}^D $
  and  $\{ \bm{\beta}^1, \dots, \bm{\beta}^{B} \} \subset \mathbb{N}^B$ for integers $P, B \ge 0$. Finally we define   $M_P := \max_{1  \le i  \le P} \| \bm{\alpha}^i \|_1$ and 
  $M_B:= \max_{1 \le i \le B} \| \bm{\beta}^i \|_1$.
% we write $Du = \{ u_{x_1}, 
% u_{x_2}, \dots, u_{x_d} \} $ to 
% denote the set of first order partial derivatives of $u$ and similarly write 
% $D^\nu u$ for $\nu > 1$ to denote the set of $\nu$-th order partial derivatives. 
Throughout the article we have in mind PDEs of the form 
\begin{subequations}\label{generic-PDE}
\begin{align}
    \mP\left( \bx, \partial^{\bm{\alpha}_1}u(\bx), \dots, \partial^{\bm{\alpha}_{P}} u(\bx) \right) & = f(\bx), \: \bx \in \Omega, \label{generic-PDE-interior} \\ 
    \mB\left(\bx, \partial^{\bm{\beta}_1} u(\bx), \dots, \partial^{\bm{\beta}_B} u(\bx) \right) & = g(\bx), \: \bx \in \partial \Omega, \label{generic-PDE-BC}
\end{align}    
% \begin{align}
%     \mP\left( \bx, u(\bx), Du(\bx), \dots, D^{M_P} u(\bx) \right) & = f(\bx), \: \bx \in \Omega, \label{generic-PDE-interior} \\ 
%     \mB\left(\bx, u(\bx), Du(\bx), \dots, D^{M_B} u(\bx) \right) & = g(\bx), \: \bx \in \partial \Omega, \label{generic-PDE-BC}
% \end{align}    
\end{subequations}
where $\mP: \R^{J_P} \to \R$  and $\mB: \R^{J_{B}} \to \R$, with $J_P= D  + P$ and $J_B = D + B$,} are
 nonlinear functions that define the functional relationships 
between {$\bx = (x_1, \dots, x_D)$} 
and values of $u$ and its partial derivatives in the interior and 
boundary of $\Omega$. 
% The dimensions  $J_P = J_P(M_P,d)$ and 
% $J_{B} = J_{B}(M_B, d)$ are simply the 
% number of partial derivatives of $u$ involved in the definition of the PDE
% and the boundary conditions. 
The functions $f: \Omega \to \R$, often 
referred to as a forcing/source term,
and $g: \partial \Omega \to \R$, the boundary condition,
constitute the data of the PDE. In most practical problems,  $M_B \le M_P$ 
and $\max\{ M_P, M_B\}$ denotes {the {\it order} of the PDE.}
{
As a running example consider the one-dimensional second order PDE 
\begin{equation}\label{darcy-1D}
\begin{aligned}
   - \partial_x \left[ a(x) \partial_x u(x) \right] + u^3(x) & = f(x), \: \: x \in (0,1),\\  
%    2x u_x(x) + (1 + x^2) u_{xx}(x) + u^3(x) & = f(x), \: x \in (0,1) \\
    u(0) = u(1) & = 0,
\end{aligned}
\end{equation}
where $a \in C^1(\overline{\Omega})$ is a spatially varying coefficient, for example, 
drawn from a random field. We assume this coefficient 
along with its first derivative can be evaluated but in general it may 
have a complicated or unknown form.
We can readily read that $\mB(x, u(x)) \equiv u(x)$.  Expanding the 
differential operator on the left hand side of the PDE 
we get 
$\mP\left(x, u(x), \partial_x u (x), \partial^2_x u(x)\right) = 
-\partial_x a(x) \partial_x u(x) - a(x) \partial^2_x u(x)  + u^3(x)$. Thus, defining the 
new variables\footnote{The entries 
$s_i, t_i$ simply denote the values of $x$ as well as  $u$ and its 
partial derivatives evaluated at $x$. This compact notation will be useful later on.}
$\mbf{s} = (s_1, \dots, s_4) \equiv [x, u(x), u_x(x), u_{xx}(x)] \in \R^4$ and 
$\mbf{t} = (t_1, t_2)  \equiv [x, u(x) ] \in \R^2$  we can write 
\begin{equation*}
\begin{aligned}
    \mP(\bs) & = -a_x(s_1)s_3 - a(s_1) s_4 + s_2^3 , \\
    \mB(\bt) & = t_2.
\end{aligned}
% \begin{aligned}
%     \mP(\bs) & = 2 s_1 s_3 + (1 + s_1^2) s_4 + s_2^3, \\
%     \mB(\bt) & = t_2,
% \end{aligned}
\end{equation*}
 Throughout the rest of the 
article we will assume that whenever a PDE is presented, it is well-defined 
and has a unique strong solution $u$, i.e., a solution that is defined pointwise. 
}

\subsection{Representer Theorems}\label{sec:representer-theorems}
Consider a simply connected set $\Theta \subseteq \R^D$.
Following \cite{muandet2017kernel},
we say that a function $\mK: \Theta \times \Theta \to \R$ is a {\it Mercer kernel} 
if it is symmetric, that is  $\mK(\bs_1, \bs_2) = \mK(\bs_2, \bs_1)$, and 
that for any collection of points $S = \{ \bs_1, \dots, \bs_n\}$ the matrix
$ \left[  \mK(S,S) \right]_{ij} = \mK(\bs_i, \bs_j)$ is positive definite. 
We write $\mH_\mK$ to denote the Reproducing Kernel Hilbert Space (RKHS) associated 
to $\mK$ with its norm denoted by $\| \cdot \|_\mK$.

Suppose {$\mH_\mK$ is continuously embedded in $C^k(\Theta)$,
the Banach space of 
real valued functions that are $k$-times continuously differentiable with $k \in \mbb{N}$ equipped with the usual sup norm},
and fix the points $X = \{ \bx_1, \dots, \bx_{J} \} \subset \Theta$.
Let $\delta_j: C^0(\Theta) \to \R$   denote the pointwise evaluation operator 
mapping a function $u \in C^0(\Theta)$ to its point value  $u(\bx_j)$  for 
$(1 \le j \le J)$,
let $L_q: C^k(\Theta) \to C^0(\Theta)$ for $q =1, \dots, Q$ 
be  bounded and linear operators, and define the maps 
\begin{equation*}
    \phi_j^q := \delta_{j} \circ L_{q}, \quad u \mapsto L_q(u) (\bx_j),  
\end{equation*}
which first apply the $L_q$ operators to {a regular function $u$} 
and then evaluate the output at the 
point $\bx_j$. For the PDE solver of \cite{chen2021solving}, these maps often evaluate
$u$ or some of its partial derivatives  at a set of 
collocation 
points.
% \zhetext{which pointwisely evaluate the operators  applied to $u$ at the collocation points. }
Concatenating the $\phi_j^q$ along the $j$ and $q$ indices we obtain 
a vector of maps $\bphi: C^\alpha(\Theta) \to \R^N$, where $N = QJ$. The 
ordering of the $\phi_j^q$ is innocuous and henceforth we write 
$\phi_n$ for $n =1, \dots, N$ to denote the entries of the vector $\bphi$.

{
We now consider optimal recovery problems of the form 
\begin{equation}\label{kernel-interpolation}
    \minimize_{u \in \mH_\mK}  \| u\|_\mK \quad \text{s.t.} \quad \bphi(u) = \bo,
\end{equation}
where $\bo \in \R^N$ is a fixed vector. 
It is well-known (see for example Chapter 12 of \cite{owhadi2019operator}) that 
the minimizer $\bar{u}$ of this optimization problem has the form
\begin{equation}\label{representer-kernel-interpolation}
    \bar{u}(\bx) = \mK(\bx, \bphi) \mK(\bphi, \bphi)^{-1} \bo.
\end{equation} 
Here $\mK(\bx, \bphi)$ is a vector field on $\Theta$ with 
entries $\left[\mK(\bx, \bphi)\right]_i = \phi_i( \mK(\bx, \cdot) )$ for $i = 1, \dots, N$ 
and $\mK(\bphi, \bphi) \in \R^{N \times N}$ is a symmetric matrix with entries 
$ \left[\mK(\bphi, \bphi) \right]_{ij} = \phi_i \otimes \phi_j (\mK)$, i.e., we apply $\phi_i$ 
along the first argument of $\mK$  then apply $\phi_j$ along the second 
argument.

One can also 
extend this formula to nonlinear regression problems of the form
\begin{equation}\label{kernel-regression}
    \minimize_{u \in \mH_\mK} \| u\|^2_\mK + \frac{1}{\lambda^2} \| \mF \circ \bphi(u) - \bo \|_2^2,
\end{equation}
where $\lambda > 0$ determines regularization strength, $\mF: \R^N \to \R^O$ is a nonlinear map, 
and $\bo \in \R^O$. 
By Proposition~2.3 of \cite{chen2021solving},
minimizers $\bar{u}$ of \eqref{kernel-regression} have the form 
\begin{equation}\label{representer-kernel-form}
    \bar{u}(\bx) = \mK(\bx, \bphi) \mK(\bphi, \bphi)^{-1} \bar{\bz},
\end{equation}
 where the vector $\bar{\bz}$ solves the optimization problem 
\begin{equation}\label{representer-optimization}
    \minimize_{\bz \in \R^N} \: \bz^T \mK(\bphi, \bphi)^{-1} \bz
    + \frac{1}{\lambda^2} \| \mF(\bz) - \bo \|_2^2.
\end{equation}
In the special case where $N = O$ and $\mF = \text{Id}$ we obtain 
\begin{equation}\label{representer-kernel-regression}
    \bar{u}(\bx) = \mK(\bx, \bphi) \left( \mK(\bphi, \bphi) + \lambda^2 I \right)^{-1} \bo,
    \end{equation}
    by analytically solving \eqref{representer-optimization} 
    and substituting in \eqref{representer-kernel-form}.
}
% If $N = O$ and  $\mF = \text{Id}$
% %@Shandian: you were correct. We need to assume $N = O$.}
% %\zhe{question: if $\mathcal{F}$ is a $R^N$ to $R^O$ mapping ($N \neq O$), why can it be an ID? }
% we can solve 
% \eqref{representer-optimization} explicitly to obtain  

% \end{equation}
% which is a familiar expression  in  kernel regression.
During the smoothing of data and PDE discovery steps of our approach 
we simply take $\mF = \text{Id}$. 
When solving nonlinear PDEs it is often the case that $O < N$ 
and  $\mF$ turns out  to be 
a nonlinear map defined via the functional form of the PDE, i.e., the functions 
$\mP, \mB$ in \eqref{generic-PDE} and so the second term in \eqref{kernel-regression} 
measures the residual of the PDE at the collocation points (see \cite{chen2021solving} for more details).

 % Citations within the text should include the author's last name and
% year, e.g., (Cheesman, 1985). 
% %Apart from including the author's last name and year, citations can follow any style, as long as the style is consistent throughout the paper.  
% Be sure that the sentence reads
% correctly if the citation is deleted: e.g., instead of ``As described
% by (Cheesman, 1985), we first frobulate the widgets,'' write ``As
% described by Cheesman (1985), we first frobulate the widgets.''

% The references listed at the end of the paper can follow any style as long as it is used consistently.

%Be sure to avoid
%accidentally disclosing author identities through citations.

\section{METHODOLOGY}\label{sec:methodology}
We now outline our kernel methodology for 
equation discovery and operator learning
of PDEs from empirical
data. {We first present an abstract three-step framework for 
equation discovery and operator learning after which we outline our kernel 
approach for each single step. }
For simplicity we will only consider the case where the function $\mP$ in \eqref{generic-PDE}
is unknown since this is most practically relevant. The approach can be 
extended to learn $\mB$ in a similar manner. We will also assume that the 
order $M_P$ of the PDE is known.

\subsection{An Abstract Framework for Equation Discovery and Operator Learning}

Suppose a set of mesh/observation points 
$X = \{\bx_j \}_{j=1}^J \subset \Omega$ is fixed and 
let $\{u^{(i)}, f^{(i)} \}_{j=1}^I$ be pairs of solutions and 
forcing terms 
for the PDE \eqref{generic-PDE} with the same boundary conditions. Our training 
data consists of noisy observations of the pairs $(u^{(i)}, f^{(i)})$ at the 
points $X$, that is, 
\begin{equation*}
\begin{aligned}
       \R^J \ni \bu^{(i)} & = u^{(i)}(X) + \bm{\eps}^{(i)},   \\ 
       \R^J \ni  \mbf{f}^{(i)} & = f^{(i)}(X),
\end{aligned}
\end{equation*}
where we used the shorthand notation $u(X) = \left( u(\bx_1), \dots, u(\bx_J) \right)$ and $\bm{\eps}^{(i)} \sim N(0, \lambda_\mU^2 I)$ is the measurement noise with 
standard deviation $\lambda_\mU >0$. Given this input/training data 
we then consider a three-step framework:

{
{\bf Step (i): Smoothing the Training Data and Estimating Derivatives.}
Consider a Banach space $\mH$ that is continuously embedded in $C^{M_p}(\Omega)$. 
Then solve  the regression  problems 
\begin{equation}\label{step-i-regression}
   \bar{u}^{(i)} = \argmin_{v \in \mH} \| v \|^r_{\mH}  
   + \frac{1}{\lambda_\mU^2} \| v(X) - \bu^{(i)} \|_2^2
   % \quad \text{s.t.} \quad 
   % v(X) + \bm{\eps}^{(i)} = \bu^{(i)},  
\end{equation}
for $i = 1, \dots, I$ and $r > 0$. Proceed to compute the partial derivatives  
$\partial^{\bm{\alpha}_j} \bar{u}^{(i)}(X)$ for $j = 1, \dots, P$, i.e., the pertinent partial derivatives of the smoothed solutions involved in \eqref{generic-PDE} evaluated  
at the mesh points $X$ \footnote{If one wishes to 
learn the boundary operator $\mB$ then the $\partial^{\bm{\beta}_j} \bar{u}^{(i)}$ should also be computed at a set of boundary 
collocation points.}.
}

{
{\bf Step (ii): Learning the Functional Form of the PDE.}
Define the set of vectors
\begin{equation}\label{formula-for-s-j}
\begin{aligned}
        \overline{\bs}_j^{(i)}
       = \left(\bx_j,  \partial^{\bm{\alpha}_1} \bar{u}^{(i)}(\bx_j), 
       \dots, \partial^{\bm{\alpha}_P} \bar{u}^{(i)}(\bx_j)\right) \in \mathbb{R}^{J_P},
\end{aligned}
\end{equation}
for $i =1, \dots, I$. Now consider another Banach space $\mH'$ 
that is continuously embedded in $C^0(\R^{J_P})$
and 
 approximate the function $\mP$ via the optimal recovery problem 
 \footnote{One can also formulate a regression problem analogous to Step (i) if 
 the $f^{(i)}(X)$ are believed to be noisy.}
\begin{equation}\label{step-ii-regression}
   \overline{\mP} = \argmin_{ \mV \in \mH'} \| \mV \|_{\mH'} \quad \text{s.t.} \quad 
   \mV(\overline{\bs}_j^{(i)}) = f^{(i)}(\bx_j),  
\end{equation}
for $i =1, \dots, I$ and $j = 1, \dots, J$.
}

{
{\bf Step (iii): Operator Learning by Solving the Learned PDE.}
Consider a new source term $\tilde{f}$ for which we wish to 
approximate 
the solution to the PDE \eqref{generic-PDE}. Since $\mP$ is unknown  
 formulate the following PDE instead
\begin{equation*}
\begin{aligned}
    \overline{\mP}\left( \bx, \partial^{\bm{\alpha}_1} u(\bx), \dots, 
    \partial^{\bm{\alpha}_P} u (\bx) \right) &= \tilde{f}(\bx), \: \bx \in \Omega, \\
    \mB \left( \bx, \partial^{\bm{\beta}_1} u(\bx), 
    \dots, \partial^{\bm{\beta}_B} u(\bx) \right) & = g(\bx), \:\bx \in \partial \Omega.
\end{aligned}
\end{equation*}
Note that $\overline{\mP}$ is the function given by \eqref{step-ii-regression} and this PDE is in general not well-posed unless we pose stringent conditions on 
$\overline{\mP}$. Henceforth we think of  ``solving" this PDE simply by 
 finding a function $\tilde{u}$ that approximately satisfies the equations. 
To do so, take new sets of collocation points 
$\{ \tilde{\bx}_1, \dots, \tilde{\bx}_{\tilde{J}_\Omega} \} \subset \Omega$ (the interior points) 
and $\{ \tilde{\bx}_{\tilde{J}_\Omega+1}, \dots, \tilde{\bx}_{\tilde{J}} \} \subset \partial \Omega$
(the boundary points). Choose parameters $r, \lambda_\mP, \lambda_\mB >0$ and 
approximate $\tilde{u}$ by solving the optimization problem 
\footnote{One can take $u$ in a different space than $\mH$ if a priori knowledge 
of its regularity exists.}
\begin{equation}\label{step-iii-regression}
\begin{aligned}
    & \hat{u} := \argmin_{u\in \mH}  \:\| u \|_{\mH}^r  \\
    & + \frac{1}{\lambda^2_\mP} \sum_{j=1}^{\tilde{J}_\Omega}  
    | \overline{\mP} \left( \tilde{\bx}_j, \partial^{\bm{\alpha}_1} u(\tilde{\bx}_j),\dots,
    \partial^{\bm{\alpha}_P} u(\tilde{\bx}_j) \right)  - \tilde{f}(\tilde{\bx}_j) |^2\\
    & + \frac{1}{\lambda^2_\mB} \hspace{-1ex}
    \sum_{j=\tilde{J}_\Omega+1}^{\tilde{J}}  \hspace{-2ex}
    | \mB \left( \tilde{\bx}_j, \partial^{\bm{\beta}_1} u(\tilde{\bx}_j), \dots, 
    \partial^{\bm{\beta}_B} u(\tilde{\bx}_j)\right)  - g(\tilde{\bx}_j) |^2.
    \end{aligned}
\end{equation}

}

\subsection{Implementation Using Kernels}\label{sec:kernel-implementation}
{
In what follows we present a kernel implementation of our three-step framework
by choosing the $\mH$ and $\mH'$ as appropriate RKHSs. 
}

%\subsubsection{Step (i)}\label{sec:kernel-smoothing}
{
{\bf Step (i):}
Let $\mU: \Omega \times \Omega \to \R$ be a Mercer kernel chosen so that its RKHS $\mH_\mU$ is continuously 
embedded in $C^{M_P}(\Omega)$. A simple choice would be the RBF kernel $\mU(\bx_1, \bx_2) = 
\exp \left( \frac{-1}{2 l_\ell^2} \| \bx_1 - \bx_2\|_2^2 \right) $
whose RKHS consists of infinitely smooth functions
\footnote{The RBF kernel 
may result in overly smoothed training data in which case  
the Mat{\'e}rn family of kernels (see \cite{genton2001classes})
may be a better choice as they 
 allow precise control over the regularity of the RKHS.}. 
Consider the regression problem \eqref{step-i-regression} with $\mH \equiv \mH_\mU$ and 
$r = 2$. 
We can solve this problem by applying formula \eqref{representer-kernel-interpolation} with
$\bphi =( \delta_1, \dots, \delta_J)$ to obtain the minimizer
\begin{equation}\label{denoised-solution}
    \bar{u}^{(i)}(\bx) = \mU(\bx, X) \left( \mU (X, X) + \lambda_\mU^2 I \right)^{-1} \bu^{(i)}.
\end{equation}
For any multi-index $\bm{\alpha}_j$ we can directly differentiate this formula to get 
\begin{equation}\label{denoised-derivative}
    \partial^{\bm{\alpha}_j} \bar{u}^{(i)}(\bx) = \partial^{\bm{\alpha}_j} \mU(\bx, X) \left( \mU (X, X) + \lambda_\mU^2 I \right)^{-1} \bu^{(i)},
\end{equation}
where we introduced the vector field 
$ \left[\partial^{\bm{\alpha}_j} \mU(\bx, X) \right]_i = \partial^{\bm{\alpha}_j} \mU(\bx, \bx_i)$, the entries of which can be computed offline using analytic expressions or 
automatic differentiation.
}

% Our first task is to perform kernel smoothing on the $\bu^{(i)}$ to filter 
% out the noise. Using \eqref{representer-kernel-regression} with 
% $\phi_j \equiv \delta_j$ we obtain the smoothed solutions 

% where 
% The noise 
% standard deviation (nugget) $\beta_\mU^2 > 0$ and the kernel 
% parameters $l_\mU$ 
%  can be tuned using hyper-parameter tuning methods such as
% cross validation (CV) or maximum likelihood (MLE).
% %Henceforth, we 
% %assume these parameters are fixed. 

% We proceed by differentiating \eqref{denoised-solution} to approximate 
% the derivatives of $u^{(i)}$, 

%\subsubsection{Step (ii)}\label{sec:PDE-learning}

{
{\bf{Step (ii):}}
With  formula \eqref{denoised-derivative} at hand we 
compute the vectors $\bar{\bs}_j^{(i)}$ following \eqref{formula-for-s-j}. 
We then choose a Mercer kernel $\mK : \R^{J_P} \times \R^{J_P} \to \R$
with RKHS $\mH_\mK$ which is assumed to be continuously embedded in $C^0(\R^{J_P})$; 
once again  the RBF kernel would be a simple choice \footnote{Another possible choice is 
the polynomial kernel; see Section~\ref{sec:experiment-details}}. We then formulate the optimal 
recovery problem \eqref{step-ii-regression} with $\mH' \equiv \mH_\mK$. Then 
by equation \eqref{representer-kernel-interpolation} we have an explicit 
formula for  $\overline{\mP}$, 
\begin{equation}\label{learned-PDE-representer-formula}
    \overline{\mP}(\bs) = \mK(\bs, S) \left( \mK(S, S) + \lambda_\mK^2 I \right)^{-1} \mbf{f},
\end{equation}
where we introduced an additional nugget parameter $\lambda_\mK >0$ to improve 
the conditioning of the kernel matrix $\mK(S, S)$ as is customary in kernel interpolation or 
regression.
}

% \begin{equation*}
% \begin{aligned}
%         \bar{\bs}_j^{(i)}
%        = \left(\bx_j,  \bar{u}^{(i)}(\bx_j), D\bar{u}^{(i)}(\bx_j), 
%    \dots, D^{M_P} \bar{u}^{(i)}(\bx_j) \right).
% \end{aligned}
% \end{equation*}
% The pairs $(\bs^{(i)}, \mbf{f}^{(i)} )$ constitute our training data for 
% equation discovery.

% Assuming that $D^\nu \bar{u}^{(i)}$ are close  to $D^\nu u$ we can 
% substitute $ \bar{u}$ in \eqref{generic-PDE-interior} and evaluate at the points  $\bx_j \in X$ to write 
% \begin{equation*}
%     \mP\left( \bs_j^{(i)} \right) = f^{(i)}(\bx_j).
% \end{equation*}
% We view these identities as interpolation constraints for the function $\mP$ 
% and use them to define a  kernel interpolant $\overline{\mP}$. 
% More precisely, let $\mK: \R^{J_P} \times \R^{J_P} \to \R$ be a 
% Mercer kernel and write $S = \{ \bs^{(1)}, \dots, \bs^{(I)} \}$ 
% and $\mbf{f} = \left( \mbf{f}^{(1)}, \dots, \mbf{f}^{(I)} \right)$. 
% We can then use \eqref{kernel-regression}  to write 
% \begin{equation}
%     \overline{\mP}(\bs) = \mK(\bs, S) \left( \mK(S, S) + \beta_\mK^2 I \right)^{-1} \mbf{f}.
% \end{equation}
% The parameter $\beta_\mK > 0$ is a nugget that  can be tuned  to 
% stabilize the inversion of the kernel matrix $\mK(S, S)$. 
% For our experiments we will take $\mK$ to be the RBF kernel 
% to have maximum flexibility and accommodate variable coefficients

% The length scale parameter $l_\mK$ of the kernel $\mK$
% can once again be tuned using standard hyper-parameter tuning methods.
% Computing $\overline{\mP}$
% constitutes the learning of the PDE.

%\subsection{Step (iii)}\label{sec:PDE-solver}

{
{\bf{Step (iii):}}
Finally we consider problem \eqref{step-iii-regression} and, following \cite{chen2021solving},
we take $\mH \equiv \mH_\mU$ and $r = 2$. Let $\tilde \delta_j$ 
denote the pointwise evaluation operator at $\tilde{x}_j$ and
define the maps $\tilde\phi_j^i = \tilde{\delta}_j \circ \partial^{\bm{\alpha}_i}$,
for $i = 1, \dots, p$ and $j =1, \dots, \tilde{J}_\Omega$ as well as 
$ \tilde\psi_j^i = \tilde{\delta}_j \circ \partial^{\bm{\beta}_i}$,
for $i = 1, \dots, q$ and $j = \tilde{J}_\Omega + 1, \dots, \tilde{J}$.
Further define the vector valued operators 
$\tilde \bphi_j := (\tilde\phi_j^1, \dots, \tilde\phi_j^p)$ and 
$\tilde \bpsi_j := (\tilde\psi_j^1, \dots, \tilde\phi_j^q)$. We can now rewrite 
\eqref{step-iii-regression} in the compact form 
\begin{equation*}
\begin{aligned}
    & \minimize_{u\in \mH_\mK}  \:\| u \|_{\mK}^2  + \frac{1}{\lambda^2_\mP} \sum_{j=1}^{\tilde{J}_\Omega}  
    \left| \overline{\mP} \left( \tilde{\bx}_j, \tilde \bphi_j(u) \right)  - \tilde{f}(\tilde{\bx}_j) \right|^2\\
    & + \frac{1}{\lambda^2_\mB}
    \sum_{j=\tilde{J}_\Omega+1}^{\tilde{J}}  
    \left| \mB \left( \tilde{\bx}_j, \tilde \bpsi_j (u) \right)  - g(\tilde{\bx}_j) \right|^2.
    \end{aligned}
\end{equation*}
Realizing that this is precisely the same form as \eqref{kernel-regression} 
we evoke formula \eqref{representer-kernel-form}
to identify the minimizer $\hat{u}$ as 
\begin{equation*}
    \hat{u}(\bx) = \mK(\bx, \tilde{\bphi}) \mK(\tilde{\bphi}, \tilde{\bphi})^{-1} \hat{\bz},
\end{equation*}
where we defined 
$\tilde{\bphi} := (\tilde \bphi_1, \dots, \tilde \bphi_{\tilde{J}_\Omega}, 
\tilde \bpsi_{\tilde J_\Omega + 1}, \dots, \tilde \bpsi_{\tilde J})$ and 
$\hat{\bz}$ is the minimizer of the optimization problem 
\begin{equation}\label{representer-OPT}
\begin{aligned}
    &\minimize_{\bz = (\bz_1, \dots, \bz_{\tilde{J}})}
    \: \bz^T \mU(\tilde{\bphi}, \tilde{\bphi})^{-1} \bz^T   + \frac{1}{\lambda^2_\mP} \sum_{j=1}^{\tilde{J}_\Omega}  
    | \overline{\mP} \left( \bz_j \right)  - \tilde{f}(\tilde{\bx}_j) |^2\\
    & + \frac{1}{\lambda^2_\mB}
    \sum_{j=\tilde{J}_\Omega+1}^{\tilde{J}}  
    | \mB \left( \bz_j \right)  - g(\tilde{\bx}_j) |^2.
\end{aligned}
\end{equation}
In practice we solve this problem  using a gradient descent algorithm, such as the 
Gauss-Newton algorithm proposed in \cite{chen2021solving} or L-BFGS.
}

\subsection{Theory}\label{sec:theory}
{
We now present a  convergence theory for the kernel 
implementation of Step (i) using classic results from theory of 
scattered data approximation indicating the desirable convergence properties of 
our implementation for denoising the training data and estimating the derivatives. 
}
{
Let us consider the idealized analgoue of problem \eqref{step-i-regression} where  
$\bm{\eps}^{(i)} \equiv 0$ leading to the optimal recovery problem 
\begin{equation}\label{step-i-regression-idealized}
   \bar{u}^{(i)} = \argmin_{v \in \mH_\mU} \| v \|_{\mU}  
   \quad \text{s.t.} \quad 
   v(X) = \bu^{(i)}.
   % \quad \text{s.t.} \quad 
   % v(X) + \bm{\eps}^{(i)} = \bu^{(i)},  
\end{equation}
Note that $X = \{\bx_j \}_{j=1}^J \subset \Omega$. We can then obtain a rate for the convergence of $\bar{u}^{(i)}$
towards $u^{(i)}$ for $i = 1, \dots, I$.
\begin{theorem}\label{thm:convergence-step-i}
Suppose $\Omega$ is simply connected, bounded, and has a Lipschitz 
boundary. Consider real numbers $k, t$ such that 
$k > t > D/2 + M_P$ and 
suppose $\mH_\mU$ is continuously embedded 
in the Sobolev space $H^k(\Omega)$ and that
$u^{(i)} \in \mH_\mU$.
Define 
the fill-distance 
 $h_{X, \Omega} := 
     \sup_{\bx' \in \Omega} \inf_{\bx \in X} | \bx - \bx' |.$ 
    Then for sufficiently small $h_{X, \Omega}$ there exists $C> 0$, independent of $u^{(i)}$ and $h_{X, \Omega}$, so that 
\begin{equation*}
    \| \bar{u}^{(i)} - u^{(i)} \|_{H^t(\Omega)} 
    \le C h_{X, \Omega}^{k- t} \| u^{(i)} \|_{\mH_\mU},
\end{equation*}
where the $\bar{u}^{(i)}$ are given by \eqref{step-i-regression-idealized}.
\end{theorem}
The proof is a direct application of  
Proposition~11.30 of \cite{wendland2004scattered} under the assumption that $\mH_\mU$ 
is embedded in the sobolev space $H^k(\Omega)$. We also note that
the above theorem does not only give a convergence rate for the 
functions $\bar{u}^{(i)}$ but also for their requisite partial derivatives of order $M_P$ thanks to the Sobolev Embedding 
theorem (See Chapter 4 of \cite{adams2003sobolev}) 
which states that $H^t(\Omega)$ is continuously embedded 
in $C^{M_P}(\Omega)$ under the hypothesis of Theorem~\ref{thm:convergence-step-i}.
}

{
One can also aim to obtain similar results
for Steps (ii) and (iii) of our approach using RKHS theory. 
Doing so is not trivial due to the propagation of errors from each step to the next. 
This analysis is the subject of the upcoming publications 
\cite{bh-1, bh-2}.
}

\section{EXPERIMENTS}\label{sec:experiments}
Below we compare our computational framework to state-of-the-art algorithms for 
equation discovery and operator learning. Here we focus on presenting the results 
and give a brief summary of the setup. Further details of experiments can
be found in Section~\ref{sec:experiment-details} of Appendix.

Three benchmark DEs 
were considered: a pendulum model \eqref{pendulum}, a nonlinear diffusion PDE \eqref{diffusionPDE}, and the Darcy flow PDE \eqref{darcyflow}. 
{For the PDE learning/discovery task we  compared our kernel 
method to  SINDy \cite{rudy2017data} for the pendulum and 
diffusion PDEs. Both our method and SINDy were trained using
the same training data with our kernel method used to
denoise the  training solutions $u^{(i)}$ and to compute the relevant partial derivatives. 
The kernel parameters in this step were tuned using CV. A test data set was then constructed by taking the same training source terms $f^{(i)}$ from the training set, 
perturbing them in a controlled manner, and solving the DEs using an independent solver. 
The Darcy flow PDE was excluded from these experiments since it is unclear 
how to choose a SINDy dictionary for PDEs with spatially variable coefficients. 
}

{
For operator learning we used our kernel method and SINDy for Steps (i) and (ii) 
and used the resulting $\overline{\mP}$'s coupled with  the kernel PDE solver 
of \cite{chen2021solving} for  Step (iii). Results were further compared with 
 the   
DeepONet algorithm \cite{lu2022comprehensive} (both the original 
version and the POD-DeepONet) and the Fourier Neural Operator (FNO) method of 
\cite{li2020fourier}, trained using the same training data set, to learn 
the mapping from the source term $f$ to the solution $u$. 
Throughout the experiments we also used a second POD-DeepONet, denoted as 
POD-DeepONet (L) in our tables, which is a large network that we 
tuned to maximize performance and achieve the closest results to our kernel method.
All operator learning methods 
were validated on a test set consisting of new pairs of solutions and source terms.
Errors were computed via comparison to an independent high-resolution PDE solver 
that was taken as ground truth.
}

\subsection{Pendulum}\label{subsec:pendulum}
The following system of ODEs modeling the motion of a pendulum was considered 
\begin{equation}\label{pendulum}
    \begin{aligned}
        (u_1)_t(t) & = u_2(t), \\
        (u_2)_t(t) & = - k \sin( u_1(t)) + f(t),
    \end{aligned}
\end{equation}
subject to $u_1(0) = u_2(0) = 0$. Note that here we used the parameter $t$ as our input parameter 
rather than just $x$  as is common notation in ODE and PDE literature.
The training data for this experiment consists of the pairs of solutions and forcing functions
$(u^{(i)}(t_j), f^{(i)}(t_j))$ for {$i =1, \dots, I$ (we took $I=10$ or $20$).}  Each forcing $f^{(i)}$ was drawn 
from a GP and the points $t_j$ were distributed uniformly; see Section~\ref{sec:experiment-details}. 
%\textcolor{green}{Da: I checked that the $t_j$ was from a grid, including the diffusion and Darcy experiments.}

% [From Da: Only RBF kernels are used. All kernel matrices but the kernel matrix for P used the nugget coefficient $\beta^2=10^{-10}$. The length scales for $\mU$ are 1.0 and 0.45 since we have two equations. The length scale 0.45 was for the $u_2$ in the second equation. The length scale for the $\mK$ is 12, and the nugget coefficient $\beta_\mK^2 = 10^{-6}$.] 
{
{\bf Equation Discovery/Learning:}
The function $\overline{\mP}$ was learned 
using our kernel approach for Step (ii) as well as SINDy; see Section~\ref{sec:pendulum-details} for details including the choice of the dictionary. 
% For each threshold, we ran 10000 times. The four thresholds had consistent results.
% \todo{add info about other hyperparameters in SINDy such as thresholding parameters etc. Scipy standard ODE solver.}
%\textcolor{green}{Da: We used the same training data from the operator learning task}. 
We took $I=20$ (size of the training set) and for testing, 
the forcing terms $f^{(i)}$  were perturbed 
using the formula $f^{(i)}_\beta = f^{(i)}(t) + \beta \sin(5 \pi t)$, the parameter $\beta$ controls 
size of the perturbation and hence, the departure of the test and training sets. 
The  ODEs were then solved using an independent solver to obtain the  
perturbed solutions $u_\beta^{(i)}$. The kernel smoothing of Step (i) was 
then used to estimate the pertinent derivatives of the $u^{(i)}_\beta$  which were then used to 
define a new set of inputs over which the error between $\overline{\mP}$ and $\mP$ was computed for our kernel method and SINDy.
The results are reported in Figure~\ref{fig:equation-learning-test} where we 
observe that our approach with $\mK$ taken to be the polynomial kernel
(see Section~\ref{sec:kernels} for the definition of our kernels)
almost perfectly 
matches SINDy (the points overlap almost perfectly) and the learned 
equations are very robust to perturbations of the test set, a sign that $\overline{\mP}$
is a good approximation to $\mP$, globally. Taking $\mK$ to be the 
ARD kernel results in different behavior where the error is larger and grows with $\beta$, 
a sign that $\overline{\mP}$  locally approximated $\mP$.
% \bf{$\mK$ taken to be the ARD kernel
% outperforms SINDy by a factor of 5. 
% The test errors for both 
% methods increase as a function of $\beta$. This is a sign that
% the function $\overline{P}$ was learned locally around the training data. \textcolor{green}{Da: The polynomial kernel had the same performance as the SINDy, while the ARD kernel was worse}.

% When the test points deviate 
% from the training set the quality of $\overline{P}$ deteriorates, a behavior that is well-understood 
% in the context of kernel regression and interpolation. 
{
{\bf Operator Learning:} 
For operator learning we used our method and SINDy to learn $\overline{\mP}$ as above 
with training data of size $I=10$ and $20$ and
 compared our three-step approach to DeepONets and FNO. The trained models were then validated on a 
test set of $50$ solution-forcing pairs that were generated by the same 
procedure as the training set. Table~\ref{tb:operator-learning-pnedulum} compares the 
average $L^2$ errors for the operator learning of the pendulum model.  
We observed that our method with the polynomial kernel is able to achieve the best 
performance although the errors are close to the ARD kernel and SINDy. 
The POD-DeepONet (L) model is the next competitive model despite being an order of 
magnitude worse and using a much larger neural network, i.e., more expensive parameterization.
}

{
We also repeated our experiments by adding Gaussian noise to the training data (we 
used a noise to signal ratio of 0.1), 
meaning that the solution-forcing terms are no longer satisfying the 
underlying DE. Results for this experiment are summarized in Table~\ref{tb:operator-learning-noisy}. As expected, this additional noise reduces the accuracy of all models 
but our method using the ARD kernel was still able to achieve the best performance. We 
note that the SINDy method also had very close performance. 
FNO achieved the next best result but it  was still worse by a factor of 2. 
}

\begin{figure}[htp]
    \centering
    \setlength{\tabcolsep}{0pt}
    \begin{tabular}[c]{cc}
    \begin{subfigure}[b]{0.24\textwidth}
    \includegraphics[width= \linewidth]{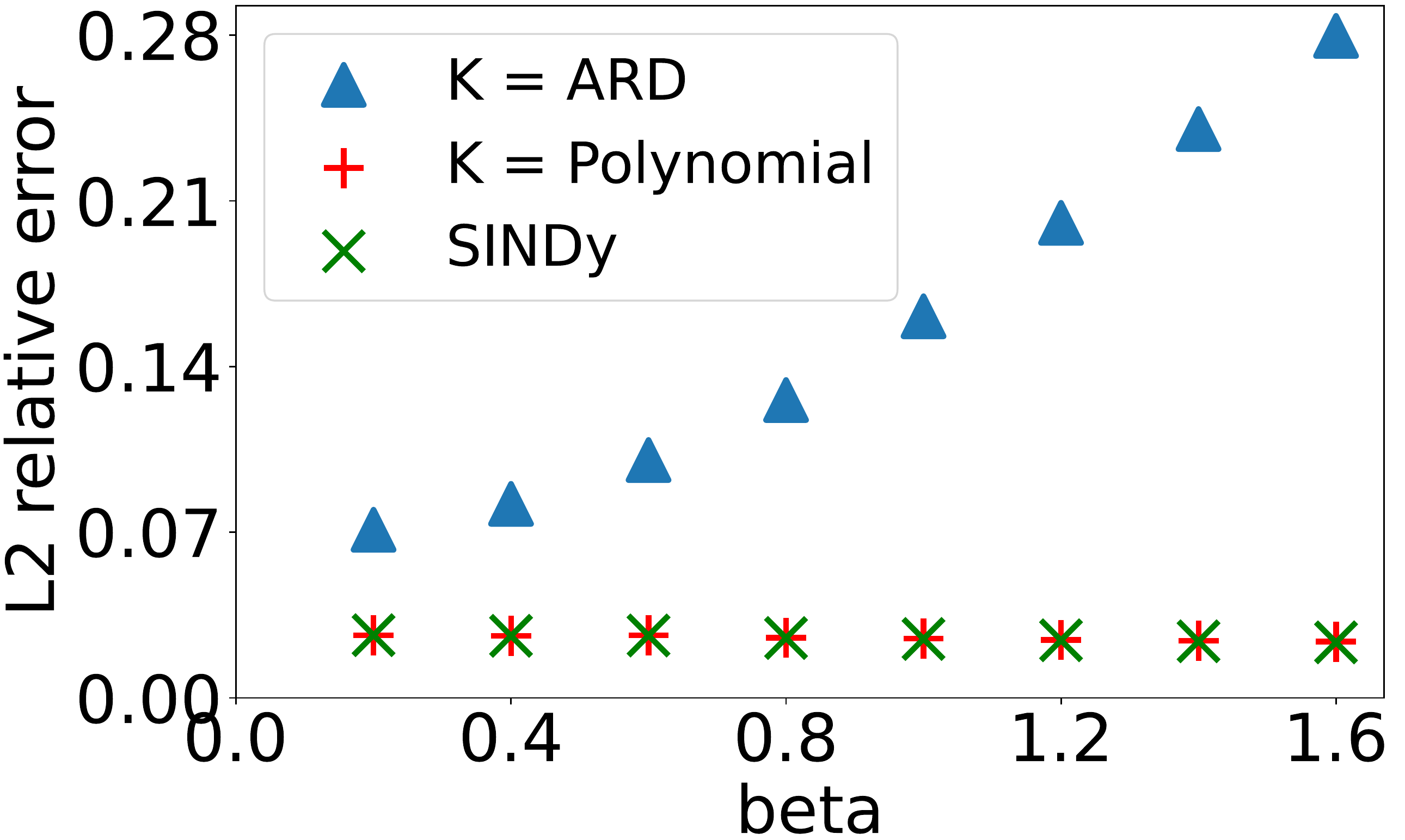}
    \end{subfigure} &
    \begin{subfigure}[b]{0.240\textwidth}
    \includegraphics[width=\linewidth]{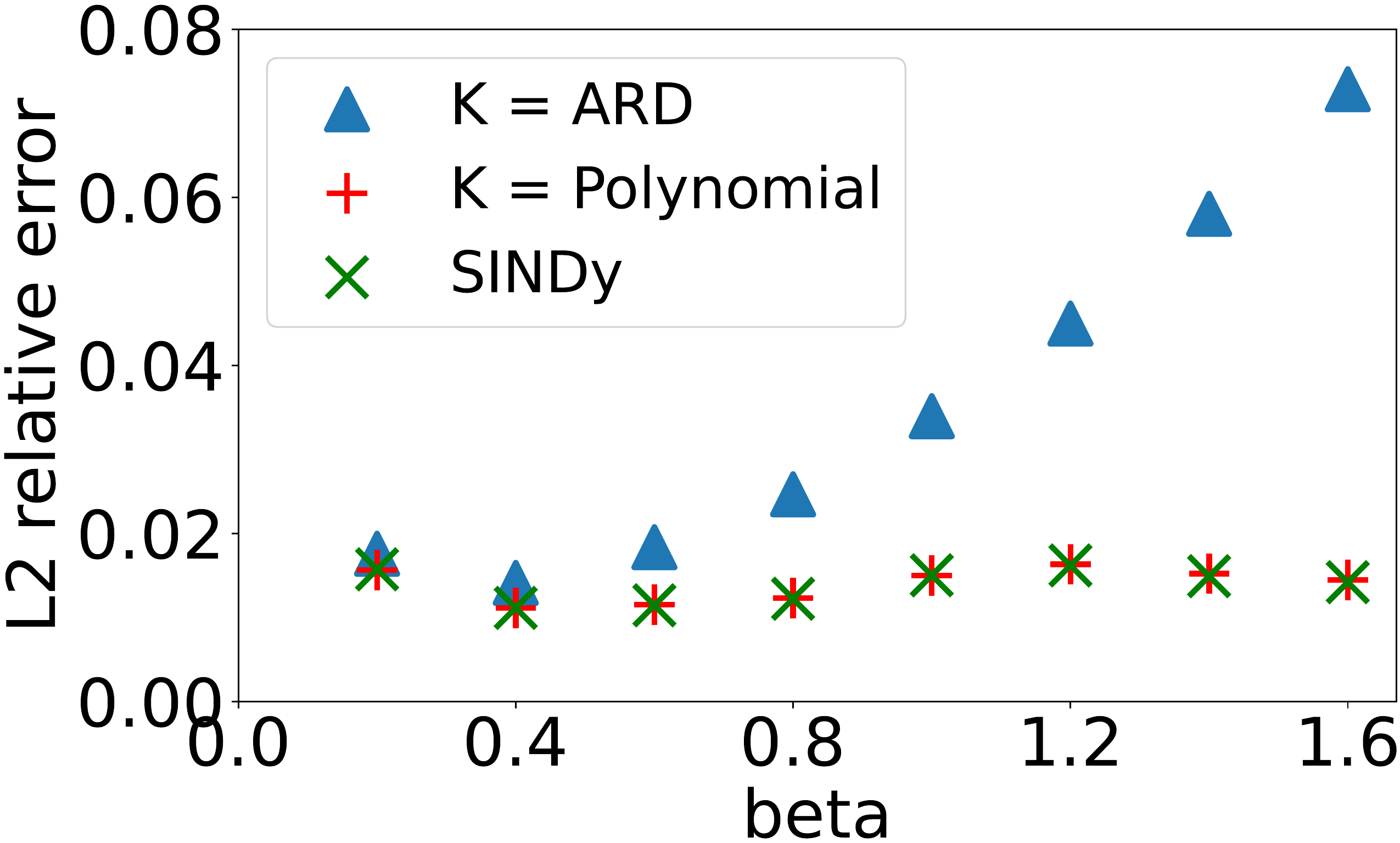}
    \end{subfigure}
    \end{tabular}
    \caption{Test error of the learned function $\overline{P}$ with our method 
    \textit{vs.} SINDy for the pendulum ODE (left) and the nonlinear diffusion PDE (right). The parameter 
    $\beta$  controls the departure of the test and training forcing terms. Results for the 
    polynomial kernel overlapped with SINDy.
    }
    \label{fig:equation-learning-test}
\end{figure}

\begin{table}[htp] 
    \centering
	\scriptsize
%	\begin{subtable}{}
		\centering
		\begin{tabular}[c]{cccc}
%			\toprule
			Method & $I =10$ & $I= 20$ \\
			\hline
			$\mK=$ ARD &  $7.5e^{-3} (1.4e^{-3})$ &  $2.3e^{-3} (3.1e^{-4})$\\
                $\mK = $ Polynomial &  $\mbf{6.3e^{-3} (1.1e^{-3})}$ &  $\mbf{2.1e^{-3} (1.7e^{-4})}$\\
            SINDy & $7.1e^{-3}(8.5e^{-4})$ & $4.5e^{-3}(4.5e^{-3})$\\
			POD-DeepONet & $1.8e^{-1}(2.7e^{-2})$ & $5.7e^{-2}(7.1e^{-3})$\\
            POD-DeepONet (L) & $3.7e^{-2}(5.3e^{-3})$ & $1.0e^{-2}(1.1e^{-3})$\\
			FNO & $1.2e^{-1}(1.3e^{-2})$ & $4.1e^{-2}(3.8e^{-3})$\\
            DeepONet & $2.9e^{-1}(2.8e^{-2})$ & $1.3e^{-1}(1.2e^{-2})$\\
%			\bottomrule
		\end{tabular}
%	\end{subtable}\\
	\caption{Average $L^{2}$ relative errors for the operator learning task of pendulum system computed for 50 test forcing functions. Standard deviations are reported in brackets. (L) indicates the large 
 network variant of POD-DeepONet. Bold text indicates the best errors.} \label{tb:operator-learning-pnedulum}
	\vspace{0.0in}
\end{table}

\begin{table}[htp] 
    \centering
	\scriptsize
%	\begin{subtable}{}
		\centering
		\begin{tabular}[c]{cccc}
%			\toprule
			Method & Pendulum & Diffusion & Darcy Flow\\
			\hline
			Our method &  $\mbf{3.9e^{-2} (2.3e^{-3})}$ &  $\mbf{6.3e^{-2} (4.6e^{-3})}$&  $7.7e^{-2} (5.0e^{-3})$\\
			POD-DeepONet & $9.7e^{-2}(1.3e^{-2})$ & $1.4e^{-1}(1.1e^{-2})$&  $9.8e^{-2} (7.2e^{-3})$\\
                POD-DeepONet (L) & $8.1e^{-2}(1.0e^{-2})$ & $1.0e^{-1}(8.8e^{-3})$& $\mbf{7.2e^{-2}(6.5e^{-3})}$\\
			FNO & $8.0e^{-2}(6.8e^{-3})$ & $7.7e^{-2}(5.0e^{-3})$& $8.8e^{-2}(9.0e^{-3})$\\
                DeepONet & $1.5e^{-1}(1.9e^{-2})$ & $2.3e^{-1}(1.8e^{-2})$& $1.5e^{-1}(1.6e^{-2})$\\
                SINDy & $4.1e^{-2}(3.8e^{-3})$ & $6.8e^{-2}(2.3e^{-3})$&N/A\\
%			\bottomrule
		\end{tabular}
%	\end{subtable}\\
	\caption{Average $L^{2}$ relative errors for the operator learning task computed for 50 test forcing functions with 0.1 noise level in the training data. Standard deviations are reported in brackets. For our method, we report the best one from the ARD kernel and the polynomial kernel.     (L) indicates the large 
 network variant of POD-DeepONet. Bold text indicates the best errors.
 } \label{tb:operator-learning-noisy}
	\vspace{0.0in}
\end{table}

% \begin{table}[t] 
%     \centering
% 	\small
% %	\begin{subtable}{}
% 		\centering
% 		\begin{tabular}[c]{cccc}
% %			\toprule
% 			Method & Pendulum & Diffusion & Darcy Flow\\
% 			\hline
% 			Our method &  $2.154e^{-5}$ &  $3.491e^{-5}$&  $9.204e^{-7}$\\
% 			SINDy & $2.021e^{-4}$ & $1.695e^{-5}$&  $N/A$\\
% 			DeepONet & $2.402e^{-2}$ & $1.742e^{-1}$&  $ 1.122e^{-5}$\\
% %			\bottomrule
% 		\end{tabular}
% %	\end{subtable}\\
% 	\caption{MSE of the learned function $\overline{P}$ compared over ... } \label{tb:equation-learning}
% 	\vspace{0.0in}
% \end{table}

\subsection{Nonlinear Diffusion PDE}\label{subsec:diffusion}
The following second order  nonlinear PDE was considered for our second set of experiments 
\begin{equation}\label{diffusionPDE}
\begin{aligned}
    u_t(x,t) = 0.01 u_{xx}(x,t)  + & 0.01 u^2(x) + f(x),  \\ 
    & (x, t) \in (0,1) \times (0,1], 
\end{aligned}
\end{equation}
subject to boundary conditions $u(0,t)  = u(1,t)  = 0$ for $t \in (0,1]$ and initial conditions
$u(x, 0)  = 0,$ for $x \in (0,1)$.
Similar to Section~\ref{subsec:pendulum}, 
the training data was generated by drawing  random sources $f^{(i)}(x)$ 
from a GP with the RBF kernel; note that  $f$ 
is only a function of $x$ here and hence a one dimensional function.
As a benchmark PDE solver in this example we used the same finite-difference solver used by 
\cite{lu2021learning}. {For detailed explanation of the setup for this experiment 
see Section~\ref{sec:diffusion-details}.}

\paragraph{Equation Discovery/Learning:}
{
We followed the same recipe as the equation discovery experiments from  Section~\ref{subsec:pendulum} to compare our kernel approach, with the 
Gaussian kernel, to SINDy. 
% \textcolor{green}{Da: We used the same training data from the operator learning task}.
We trained the model using a training data set of size $I=20$ and 
tested the learned $\overline{\mP}$ functions on a test set that 
was obtained via perturbation of the training set, parameterized by the 
$\beta$ parameter. The results of our experiments are presented in Figure~\ref{fig:equation-learning-test}. Here we see a similar picture to the case of the pendulum ODE, i.e, the 
polynomial kernel matched the performance of SINDy, and yielded a global approximation 
while the ARD kernel resulted in a local approximant.
% Interestingly, we see that SINDy outperforms our method. 
% We make two important observations: First, the SINDy error is lower than ours and second, the performance of SINDy appears to be 
% robust to noise and independent of the $\beta$ parameter while our error increases 
% with $\beta$, akin to the pendulum example. We attribute this behavior 
% to the choice of a good dictionary for SINDy as well as the enfored 
% sparsity of that model; see Section~\ref{sec:discussion} for more discussion. 
}

{
\paragraph{Operator Learning:}
For operator learning experiments we followed the recipe of Section~\ref{subsec:pendulum}
once more. All models were trained on data sets of size $I =10$ and $20$ and 
validated on a test set of size $50$, all generated using the same 
procedure but independently. Table~\ref{tb:operator-learning-diffusion} 
summarizes our results with the exact training data.  We 
observe similar trends as the pendulum example with the polynomial kernel achieving 
the best errors with SINDy achieving slightly worse performance. Interestingly, 
in this case the ARD kernel appears to perform significantly worse. Among 
the neural net methods the large POD-DeepONet  was most competitive. 
We also performed the experiments after adding artificial noise to the 
training data; the results are presented in Table~\ref{tb:operator-learning-noisy}. 
Once again we found that our method achieved the lowest error, followed closely by 
SINDy. The FNO was once again the best performing neural net based method.
}

\begin{table}[htp] 
    \centering
	\scriptsize
%	\begin{subtable}{}
		\centering
		\begin{tabular}[c]{cccc}
%			\toprule
			Method & $I=10$  & $I = 20$  \\
			\hline
			$\mK=$ ARD &  $1.3e^{-2} (2.1e^{-3})$ &  $7.5e^{-3} (1.1e^{-3})$\\
                $\mK=$ Polynomial &  $\mbf{7.0e^{-3} (5.7e^{-4})}$ &  $\mbf{4.1e^{-3} (2.4e^{-4})}$\\
			POD-DeepONet & $1.7e^{-1}(1.5e^{-2})$ & $7.8e^{-2}(6.1e^{-3})$\\
                POD-DeepONet (L) & $4.4e^{-2}(3.6e^{-3})$ & $1.4e^{-2}(1.3e^{-3})$\\
			FNO & $5.8e^{-2}(4.1e^{-3})$ & $1.6e^{-2}(1.3e^{-3})$\\
                DeepONet & $3.4e^{-1}(1.9e^{-2})$ & $1.8e^{-1}(1.5e^{-2})$\\
                SINDy & $9.6e^{-3}(9.3e^{-4})$ & $4.2e^{-3}(2.3e^{-4})$\\
%			\bottomrule
		\end{tabular}
%	\end{subtable}\\
	\caption{Average $L^{2}$ relative errors for the operator learning task of diffusion computed for 50 test forcing functions. Standard deviations are reported in brackets. 
 (L) indicates the large 
 network variant of POD-DeepONet. Bold text indicates the best errors.
 } \label{tb:operator-learning-diffusion}
	\vspace{0.0in}
\end{table}

%\subsection{Burgers' Equation}\label{subsec:Burgers}

\subsection{Darcy Flow}\label{subsec:Darcy}
For our third and final example we considered the Darcy flow PDE 
\begin{equation}\label{darcyflow}
    - \text{div} \left(  a \nabla u  \right) (x) = f(x), \: x \in (0,1)^2,   
\end{equation}
subject to homogeneous Dirichlet boundary conditions. The coefficient $a$ is 
a spatially variable field given by 
    $a(x) = \exp \left( \sin(  \pi x_1) + \sin(  \pi x_2) \right)  
     + \exp \left( - \sin (  \pi x_1) - \sin(  \pi x_2) \right).$
In this experiment we excluded SINDy as 
the construction of 
an appropriate dictionary for PDEs with spatially variable coefficients is 
\textit{not} possible without prior knowledge of the form of the 
PDE; see Section~\ref{sec:discussion}.  Therefore, here we  
focus primarily on the operator learning problem and compare our method with 
the neural net based approaches.

{
Our experiments follow a similar setup to the previous problems; see Section~\ref{seC:darcy-details} for details. Once again the models were trained using data sets of size
$I=10$ or $20$ and tested on a set of size $50$ with forcing terms drawn from a GP. We 
also excluded the polynomial kernel as it was not competitive in this example. 
The results of our experiments with exact training data are summarized in Table~\ref{tb:operator-learning-darcy} where our method with the ARD kernel 
achieved the lowest error followed closely by the large POD-DeepONet. Experimental 
results with the noisy training set are presented in Table~\ref{tb:operator-learning-noisy}.
Interestingly, in this setting large POD-DeepONet achieved the best errors 
followed very closely by our method (the difference is well withing the 
standard deviation of the errors). In fact, the difference between our method, 
POD-DeepONet and FNO was quite small in this experiment compared to the previous 
two examples. 
}

\begin{table}[htp] 
    \centering
	\scriptsize
%	\begin{subtable}{}
		\centering
		\begin{tabular}[c]{cccc}
%			\toprule
			Method & 10 sources & 20 sources \\
			\hline
			$\mK= $ARD  &  $\mbf{1.4e^{-2} (1.5e^{-3})}$ &  $\mbf{7.1e^{-3} (1.0e^{-3})}$\\
			POD-DeepONet & $1.1e^{-1}(1.2e^{-2})$ & $3.6e^{-2}(3.2e^{-3})$\\
                POD-DeepONet (L) & $1.7e^{-2}(1.6e^{-3})$ & $1.1e^{-2}(1.1e^{-3})$\\
			FNO & $2.3e^{-1}(2.3e^{-2})$ & $4.3e^{-2}(3.6e^{-3})$\\
                DeepONet & $3.7e^{-1}(4.2e^{-2})$ & $1.2e^{-1}(1.4e^{-2})$\\
%                SINDy & $N/A$ & $N/A$\\
%			\bottomrule
		\end{tabular}
%	\end{subtable}\\
	\caption{Average $L^{2}$ relative errors for the operator learning task of Darcy Flow computed for 50 test forcing functions. Standard deviations are reported in brackets.
  (L) indicates the large 
 network variant of POD-DeepONet. Bold text indicates the best errors.
 } \label{tb:operator-learning-darcy}
	\vspace{0.0in}
\end{table}

\section{Discussion}\label{sec:discussion}

Here we present a discussion regarding the various aspects of our method,
how it compares to SINDy as well as neural net operator learning methods, and the implications 
of our experiments.

{
{\bf Main Takeaways From Experiments:}
Our experiments 
focused on the two distinct  tasks of 
equation discovery and  operator learning. Our results concerning equation discovery 
led to three primary observations: (a) kernel smoothing is a good pre-processing step 
for denoising and estimation of gradient information before learning DEs for both 
SINDy and our approach. In fact, our kernel method for Step (i) 
extends the applicability of SINDy to training data that is provided on unstructured meshes. 
(b) the performance of our method is closely tied to the choice of the 
kernel $\mK$. With the polynomial kernel we matched the performance of SINDy 
in the pendulum and diffusion examples while the ARD kernel resulted in a local 
approximation to $\mP$; (c) Our kernel approach is more widely applicable than SINDy 
as demonstrated with the Darcy flow PDE where it is unclear how one could construct a 
dictionary for SINDy to begin with. Interestingly, here the ARD kernel appeared to 
yield good results while the polynomial kernel was far from being competitive.
}

{
Our results concerning operator learning led to two primary observations: (a) 
operator learning via PDE discovery consistently outperformed neural net based 
methods when exact training data was available (often by an order of magnitude).
(b) the performance gap was smaller when noisy training data was involved but even 
then our method was (barely) beaten by the POD-DeepONet algorithm for the Darcy Flow example only. It is noteworthy that the POD-DeepONet was using a 
significantly larger set of parameters than our (much simpler) kernel method and 
it took significant tuning and architecture adjustment to achieve this level 
of performance. 
}
% In the regime of small training data, operator learning via equation discovery outperforms more general methods 
% such as DeepONets that learn high-dimensional maps between function spaces. This is not surprising 
% since our approach uses explicit knowledge of the fact that the map of interest is the 
% solution operator of a differential equation. We also note that this approach is possible thanks 
% to the solver of \cite{chen2021solving} since equation discovery algorithms can 
% easily lead to ill-posed or unstable equations that cannot be solved by conventional solvers. 
% (b) Algorithms that are more accurate and robust at the equation 
% discovery step also perform better for operator learning independent of the employed solver.

{
{\bf Our Abstract Framework:} We highlight that our abstract three-step framework from Section~\ref{sec:methodology} encompasses many  existing 
 equation discovery methods and extends them to perform operator learning. For example, 
 choosing $\mH$ in Step (i) to be the appropriate RKHS associated to splines, we 
 obtain the spline method implemented in the PySINDy package \cite{de2020pysindy} for estimating 
 gradients.
 One can also take $\mH$ to be a Barron space \cite{ma2022barron} to obtain 
 a neural net approximation for the derivatives. Choosing a 
 sparsity promoting norm such as a $0$-norm or a $1$-norm (with $r=1$) in Step (ii) 
 yields the SINDy framework \cite{brunton2016discovering} and its relatives
 while a Barron norm will once again yield a neural net approach such as 
 PDE-Net \cite{long2018pde}. The same is also true for Step (iii), 
 one can choose $\mH$ to be a nerual net space to obtain solvers 
 such as PINNs \cite{raissi2019physics}, or a discretized Sobolev space 
 to obtain a classic finite element method, or an RKHS to obtain the 
 approach of \cite{chen2021solving}.  
 }

{
{\bf Discovering PDEs and the Role of Sparsity:} 
The primary focus of the PDE/equation discovery literature (see for example
\cite{bongard2007automated, schmidt2009distilling, brunton2016discovering, schaeffer2017learning}) has been the extraction of explicit and 
interpretable equations that describe natural laws that govern 
physical processes. In our framework, this amounts to finding a simple 
and elegant expression for $\overline{\mP}$. It is therefore natural 
to formulate Step (ii) over an appropriate 
set of features for $\overline{\mP}$ and impose a sparsity 
assumption on those features, amounting to approaches such as SINDy. 
Our kernel approach on the other hand, does not aim to find a 
human interpretable expression for $\overline{\mP}$ but rather 
approximates 
$\mP$ with a large number of features (possibly infinite) with the hope of 
achieving the best pointwise approximation error. Whether or not one chooses to 
employ the sparse dictionary/feature map approach or our kernel method, 
should be decided by downstream tasks that one wishes to perform with $\overline{\mP}$. If we wish to discover a new interpretable natural law, then sparsity
promoting methods are appropriate. If the goal is to perform operator 
learning, or if we are not confident in the quality of the dictionary,  then the kernel method is more suitable. 
}

{
{\bf Choosing Dictionaries and Learning kernels:}
It is well-known that the performance of sparsity promoting 
 methods such as SINDy is closely tied to the construction of 
 a good dictionary, in fact, in all of our experiments we used the 
 dictionaries that were suggested by previous authors and were known to give 
 competitive results. Put simply, if we have a good dictionary and 
the training data is sufficient, then we expect sparsity promoting 
methods to perform well. This fact has motivated various 
approaches, such as the Ensemble-SINDy \cite{fasel2022ensemble}, 
that aim to automate and improve the construction of dictionaries. 
However, there are various situations where the explicit construction 
of a dictionary is impossible. Consider our Darcy flow PDE 
\eqref{darcy-1D} with a coefficient $a(x)$ that is unknown. 
In this case one cannot construct a simple dictionary of functions (such as 
polynomials) of the input variables to $\mP$. This is of course
possible if we knew the regularity of $a(x)$  and the manner 
in which $\mP$ depends on $a$ but then we are injecting strong prior 
information into the problem. 

Broadly speaking, the kernel approach, 
thanks to its large/infinite number of feature maps is more suitable 
in situations where very little information about the form of $\mP$
is available and variable coefficients exist. Another major advantage of 
the kernel approach is that it naturally accommodates the tuning/learning 
of kernel parameters which amounts to tailoring the feature maps to 
the problem at hand. This extra flexibility is what allowed us to obtain 
superior results in our operator learning experiments using CV.
}

{
{\bf Operator Learning via PDE Discovery vs Direct Regression:}
At the moment the dominant approach to operator learning in the literature 
can be broadly categorized as regression of maps between function spaces. 
Many existing algorithms such as DeepONets \cite{lu2021learning, lu2022comprehensive}, 
FNOs \cite{li2020fourier, anandkumar2020neural}, 
the multipole graph neural operator \cite{li2020multipole}, 
and the PCA-Net \cite{bhattacharya2021model}, fall within this category. 
Our approach to operator learning is fundamentally different 
from these methods as it relies on first learning the 
functional form of the PDE (that is $\mP$), and then solving the 
learned PDE with a new forcing or boundary data. To our knowledge, our 
approach is the first of its kind and our experiments 
suggest that operator learning via PDE discovery is significantly more 
data efficient and gives superior performance. We conjecture this is due to the fact that our 
approach is tailored to PDEs and makes explicit use of our prior knowledge 
that the operator at hand is the solution map of a PDE, while 
the aforementioned techniques are more general.
}

\section{CONCLUSION}\label{sec:conclusion}

{
An abstract three-step computational framework was presented  for 
the discovery of DEs and 
operator learning of their solution maps via PDE discovery. A novel 
kernel implementation of this framework was presented 
and  compared with state-of-the-art algorithms. 
Our experiments demonstrated that our method for equation discovery 
is competitively accurate and robust to noise while remaining applicable 
in broad such as PDEs with variable coefficients. For operator learning 
our method is significantly more data efficient when compared to general 
neural net  methods that learn mappings between function spaces. 
}

\bibliographystyle{icml2022}
\bibliography{GoEqDiscovery-Refs}

\section*{Appendix}
\appendix

\section{Review of Relevant Literature}\label{sec:related-lit}

Below we present a review of the relevant literature to our work 
focusing on discovering/learning of PDEs, Operator 
Learning, and PDE solvers that use GPs and Kernels.

 \subsection{Discovering PDEs}
 
 Identifying the parameters of a differential equation (DE) is a well-known 
 inverse problem;
 see the works of \cite{bock1983recent, bock1981numerical}
on parameter identification of ordinary differential equations (ODEs) as well as 
the book of \cite{kaipio2006statistical} and the article of 
\cite{stuart2010inverse} for examples involving PDEs. Such problems 
are also encountered in optimal control of PDEs as outlined in the book of 
\cite{troltzsch2010optimal}. However, these classic approaches operate 
under the assumption that the expression of the DE is known up to 
free parameters that need to be identified from experimental data. 
Indeed, the approach of 
\cite{chen2021solving}  readily extends 
to solving such inverse problems.

Equation discovery/learning is a more recent problem attributed
to \cite{bongard2007automated, schmidt2009distilling} 
who used symbolic regression to discover underlying physical laws from 
experimental data. Compared with the aforementioned inverse problems, the goal here is   to discover the very form of the DE as well as  its 
parameters from experimental data. DEs that describe real world physical systems 
involve only a few terms and often have simple expressions. Based on 
this philosophy, recent approaches to equation learning try to learn a DE 
(i.e., the function $\mP$ in our formulation)
from a dictionary of possible terms/features 
along with a sparsity assumption to ensure 
only a few terms will be active. Perhaps the best known example of 
such an approach is the SINDy algorithm of \cite{brunton2016discovering, rudy2017data}. SINDy has 
 been expanded in many directions ever since (see \cite{de2020pysindy} 
 and references within) and other authors 
  have considered similar approaches \cite{schaeffer2017learning, kang2021ident}. At a high level,
 the differences between these approaches are in  the formulation of the 
 symbolic regression problem and the implementation of a sparsity assumption 
 on the features ($L_1, L_0$ regularization or various thresholding methods) as well as how they deal with noise in the training data. 

 Compared to the feature map perspective of SINDy-type methods, our approach employs a
 kernel perspective towards learning 
 $\mP$. As a result, we give up the immediate interpretability of the learned function $\overline{\mP}$ in favor of more features and a more convenient computational framework that is also able to deal with 
 more general PDEs, such as those involving spatial or temporally varying parameters. Feature based methods often cannot deal with such problems 
 since the construction of appropriate features may require prior knowledge of the general form of the
 variable coefficients. Our method can also be combined with the 
 kernel mode decomposition approach of \cite{owhadi2021kernel} to extract the dominant features in the learned function $\overline{\mP}$, thereby 
 making our approach more interpretable. 
 Additionally, our method opens the door for analyzing the accuracy and robustness of the estimator $\overline{\mP}$. 
 Such theoretical questions have 
 attracted attention very recently \cite{he2022asymptotic, he2022much} although many open questions remain. 
 Another closely related approach to our framework is the PDE-Net 
 of \cite{long2018pde, long2019pde} which, put simply, parameterizes $\overline{\mP}$ via a convolutional neural network. In this sense 
 the PDE-Net approach can also be cast within our three step 
 abstract framework by replacing the space $\mH'$ in Step(ii) 
 with a neural network space, ex. a Barron space \cite{ma2022barron}.

\subsection{Operator Learning}
Approximation or learning of the solution maps of PDEs is a vast 
area of research in applied mathematics and engineering. 
In the setting of stochastic and parametric PDEs, the 
goal is often to approximate the solution of a PDE 
as a function of a random or uncertain parameter. 
The well-established approach to such problems is 
to pick or find appropriate bases for the input parameter 
and the solution of the PDE and then  
construct a parametric, high-dimensional map, that 
transforms the input basis coefficients to the output 
coefficients. Well-established methods such as 
polynomial chaos, stochastic finite element methods, reduced basis methods, and reduced order models  \cite{ghanem2003stochastic, xiu2010numerical, cohen2015approximation, 
hesthaven2016certified, lucia2004reduced} 
fall within this category. A vast literature in applied mathematics exists on this subject and the theoretical 
analysis of these methods has been the subject of extensive 
research; see for example \cite{beck2012optimal, Chkifa2012, chkifa2014high, nobile2008sparse, nobile2008anisotropic, gunzburger2014stochastic}. More recent neural net 
based methods such as DeepONets \cite{lu2021learning}, 
and FNO \cite{li2020fourier, bhattacharya2021model, anandkumar2020neural} also fall within the aforementioned 
category of methods where the main novelty appears to be 
the use of novel neural network architectures that are very 
flexible, expressive, and allow the algorithm to learn and adapt 
the bases that are selected for the input and outputs of the 
solution map. 

In contrast to the aforementioned methods, our three-step framework 
takes a different path towards operator learning. First, we 
formulate a regression problem in Step (ii) that approximates the algebraic form of the PDE, which is a much easier problem than direct approximation of the solution map. We then approximately evaluate the solution map by solving an optimization problem that 
solves a "nearby" PDE. To this end, our method is making explicit 
use of the knowledge that the operator of interest is the solution map of a PDE. One can also unroll the steps of the optimization 
problems in steps (ii) and (iii) of our method to obtain a neural net architecture for operator learning but it is unclear if this 
direction will lead to a reasonable algorithm.

\subsection{Solving PDEs with GPs and Kernel methods}

Finally, we mention that the key to our operator learning 
framework is the existence of flexible, meshless, and general 
purpose nonlinear PDE solvers such as the kernel method of \cite{chen2021solving} or PINNs \cite{raissi2019physics}
that allow us to "solve" PDEs of the form $\overline{\mP}(u)= f$, 
which in general, are ill-defined and ill-posed. When our 
kernel approach to Step (ii) is employed we may 
end up with a function $\overline{\mP}$ that has 
infinitely many feature maps, then a  finite difference or 
finite element discretization of the resulting PDE is 
impossible or at least very expensive. This issue persists if 
we use a method such as SINDy since, even if $\overline{\mP}$
has few terms, it may still involve stiff or unstable terms 
that need specialized solvers even if the true PDE is well-behaved.
The above mentioned solvers, especially the kernel method of \cite{chen2021solving}, allow us to overcome this difficulty 
since the solution of the equation is naturally regularized 
via the RKHS norm penalty. We emphasize that this penalty 
only provides stability at this stage and does not guarantee 
that the computed solution is actually accurate. 

We also note that the use of kernel methods and GPs for solving 
PDEs has been an active area of research over the last 
decade; see for example \cite{jidling2017linearly, schmidt2021probabilistic, gulian2022gaussian, zhang2022augmented, kramer2022probabilistic, besginowconstraining}. 
Although the overwhelming majority of the research in this direction 
appears to be focused on the case of linear DEs. Some  notable 
exceptions are \cite{raissi2017machine, chen2021solving, mou2022numerical}.

\section{Details of Experiments}\label{sec:experiment-details}

Below we present additional details regarding our experiments in 
Section~\ref{sec:experiments}.

\subsection{Common Setup}

For the kernel PDE solver in Step (iii) 
we used the implementation of \cite{chen2021solving}
(\url{https://github.com/yifanc96/NonLinPDEs-GPsolver}).
For estimation of 
derivatives in our method and the training of DeepONets we used Jax. 
For FNO we used the code base provided by the authors in \cite{li2020fourier}.
The POD-DeepONet was implemented using Pytorch.
We used Python to implement SINDy, with iterative thresholding, with NumPy for the least squares step.
% Our method and DeepONet were implemented by Jax. We used the FNO codebase provided by the authors. The POD-DeepONet was implemented by Pytorch.

For all three DEs we conducted the experiments with $I=10$ and $20$ pairs of  solutions-sources (the 
$u^{(i)}, f^{(i)}$ pairs in Section~\ref{sec:methodology} )
in the training set. In the $I=20$ case we also conducted experiments with a noisy training set 
where a Gaussian noise of noise-to-signal ratio 0.1 was added to both 
the training solutions and the training sources. In all of these experiments we 
validated the models on the same test set of 50 solution-source pairs.

For solving the optimization problem \eqref{representer-OPT}
 we used the Gaussian-Newton algorithm of \cite{chen2021solving} 
 for the pendulum ODE and the Diffusion PDE with 50 iterations. In the case 
 of the Darcy flow PDE we ran  4000 steps of LBFGS  with step sizes of 0.2 and 0.5. For all of the kernel 
 matrices involved in our implementation we used diagonal nugget terms of the 
 form $\lambda I$, where $\lambda > 0$ is a constant and $I$ is an identity matrix of 
 the same size as the requisite kernel matrix; also see Section~\ref{sec:kernel-implementation}. The value of $\eta$ was tuned 
 for each experiment separately; see tables~\ref{tb:hyperpar-pendulum}--\ref{tb:hyperpar-Darcy}
 for a summary of the chosen nuggets.

For the POD-DeepONet we set  the number of bases to maximum and varied the number of hidden layers from 2 to 3, and the width from 256, 512, and 1024. We trained for 100000 epochs to ensure convergence.
We also trained a large variant of the POD-DeepONet (denoted as POD-DeepONet (L)) in Tables~\ref{tb:operator-learning-pnedulum}--\ref{tb:operator-learning-darcy}, where we set 
the width of the network to $8192$.
We implemented  FNO  using the standard four layer architecture for the integral operators, and varied the width over 64, 128, and 256. We trained the model for 4000 epochs to make sure it had 
converged. Finally we implemented  the standard DeepONet with 2 and 3 hidden layers and  varied the width from 256, 512, and 1024 and trained for 100000 epochs. All of the above 
neural nets were trained using the the Adam optimizer. We also used different activation functions (GELU, Tanh, and ReLU) and  varied the learning rates from $1e^{-3}$, $1e^{-4}$, and $1e^{-5}$.
The results reported in the Tables~\ref{tb:operator-learning-pnedulum}--\ref{tb:operator-learning-darcy} for each neural net method were the best test errors that were 
obtained by searching over the aforementioned set of 
architectures and hyperparameters.

%  For testing POD-DeepONet, we set the number of bases to maximum, varied the number of hidden layers from 2 to 3, and varied the width from 256, 512, and 1024. We ran it 100000 epochs to make sure it converges. 
%  to make sure it converges. We noticed that the POD-DeepONet was likely to outperform our method as more parameters were used, so we ran it with width 8192 and reported the results.  For testing the performance with noise involved, we added 0.1 noise to both the observations of training solutions and the training sources. }
% }

\subsection{The Kernels}\label{sec:kernels}

Throughout our experiments we used three kernels in Steps (i)--(iii) of our framework. 
The Gaussian kernel (also known as the squared exponential kernel)
\begin{equation*}
 \mK_{\text{Gauss}}(\bt, \bt') = \exp\left( - \frac{\| \bt - \bt' \|^2 }{2 \sigma^2} \right) \quad \bt \in \R^D,
\end{equation*}
with hyper-parameter $\sigma >0$. We primarily used this kernel in Step (i) 
of all of our experiments 
for smoothing the training solutions $u^{(i)}$ and estimating their requisite partial derivatives. 
The same kernel was also used in Step (iii) and during the implementation of the kernel solver 
of \cite{chen2021solving}.

We also considered a tensorized version of this kernel, which we referred to as the 
(automatic relevance determination) ARD  kernel 
in our experiments: 
\begin{equation*}
 \mK_{\text{ARD}}(\bt, \bt') = \prod_{j=1}^D \exp\left( - \frac{| t_j - t_j' |^2 }{2 \ell_j^2} \right)
 \quad \bt \in \R^D,
\end{equation*}
with hyper-parameters $\ell_1, \dots \ell_D >0$. The ARD kernel is simply a tensorization of 
1D Gaussian kernels which uses a different length scale along each input coordinate. 
Finally, we also used the polynomial kernel 
\begin{equation*}
 \mK_{\text{Poly}}(\bt, \bt') = (\bt^T \bt' + c )^d,
 \quad \bt \in \R^D,
\end{equation*}
with hyper-parameters $c >0$ and $d \in \mathbb{N}$. We only considered $d = 2, 3, 4,$ and $5$. 
For all experiments we used cross validation to choose the hyperparameters.
The ARD and polynomial kernels were used in Step (ii) of our framework.

\subsection{Pendulum}\label{sec:pendulum-details}

The training data was generated by the following recipe: the 
source terms $f^{(i)}$ for $i=1, \dots, I$ were drawn independently
from a GP with the RBF kernel and lengthscale 0.2, 
For each source term the ODE was solved using the SciPy solve\_ivp function  on a fine grid and sub-sampled over a uniform grid of the $t_j$'s for $j=1, \dots, 30$.
The test data was generated using the same recipe except that $50$ independent source terms 
were drawn. For operator learning the $L^2$ errors between the predicted solutions 
and the test solutions were computed over the $t_j$ grid and then averaged over the test set, 
the values reported in Tables~\ref{tb:operator-learning-pnedulum}--\ref{tb:operator-learning-darcy}.
When implementing SINDy, we implemented the first equation exactly and only learned 
the second equation using the dictionary 
 $\{(u_2)_t(t), u_1(t),
u_1(t)^2, u_1(t)^3, \sin(u_1(t)), \cos(u_1(t)) ,1\}$. 

When implementing our method we learned each equation in the system separately 
assuming that the right hand side for each coordinate is a function of both $u_1$ and $u_2$, 
i.e., we considered the system of ODEs 
\begin{equation*}
\begin{aligned}
    (u_1)_t(t) & = \overline{\mP}_1\left( u_1(t),  u_2(t) \right) \\
    (u_2)_t(t) & = \overline{\mP}_2\left( u_1(t),  u_2(t) \right).
\end{aligned}
\end{equation*}
All hyperparameters involved in the 
training of our kernel method for this example are summarized in 
Table~\ref{tb:hyperpar-pendulum}. We used the Gaussian kernel for Step (i)  but length scales 
were tuned for each instance of the data separately, therefore we report only the range of 
$\sigma$ for each coordinate of the solution. The Gaussian kernel was also used for Step (iii)
with a lenghscale that was chosen in the same range that was tuned for Step (i). We also used different lengthscales for 
each of $\overline{\mP}_1$ and $\overline{\mP}_2$ as indicated in the table. 

\begin{figure}
    \centering
    \setlength{\tabcolsep}{0pt}
    % \begin{tabular}[c]{ccc}
    \begin{subfigure}[b]{0.30\textwidth}
    \includegraphics[width= \linewidth]{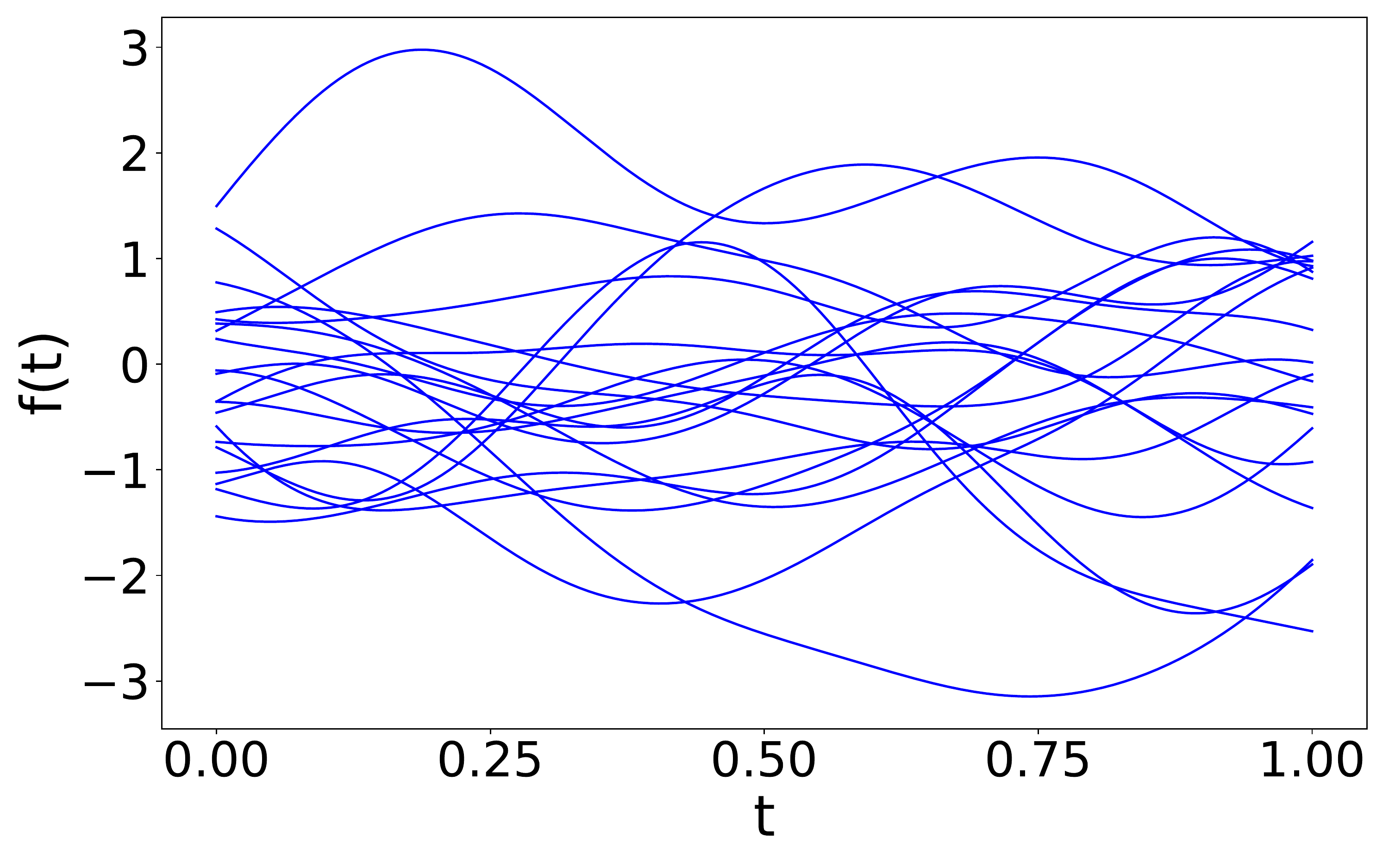} 
    \end{subfigure} 
    \begin{subfigure}[b]{0.30\textwidth}
    \includegraphics[width= \linewidth]{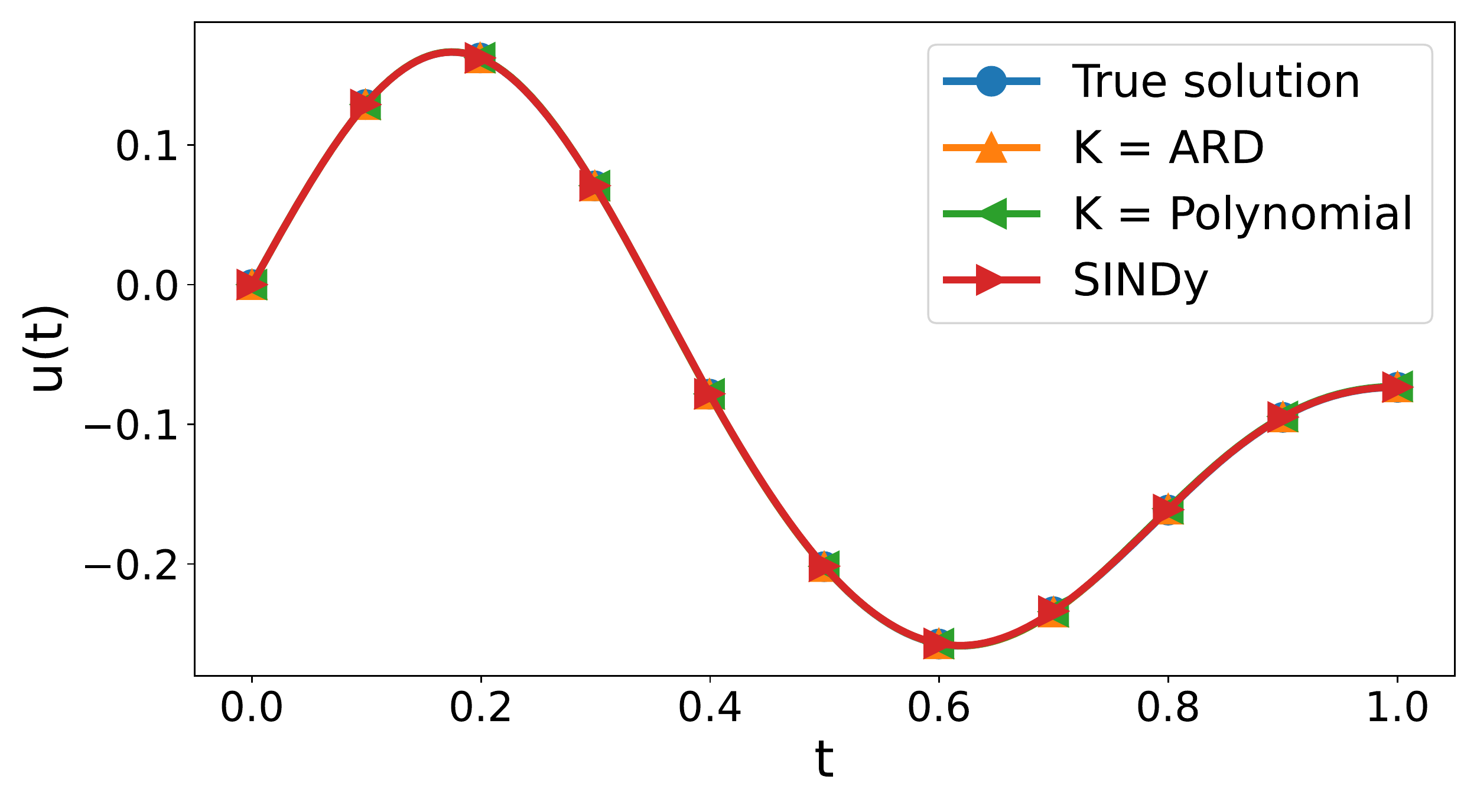}
    \end{subfigure} 
    \begin{subfigure}[b]{0.30\textwidth}
    \includegraphics[width= \linewidth]{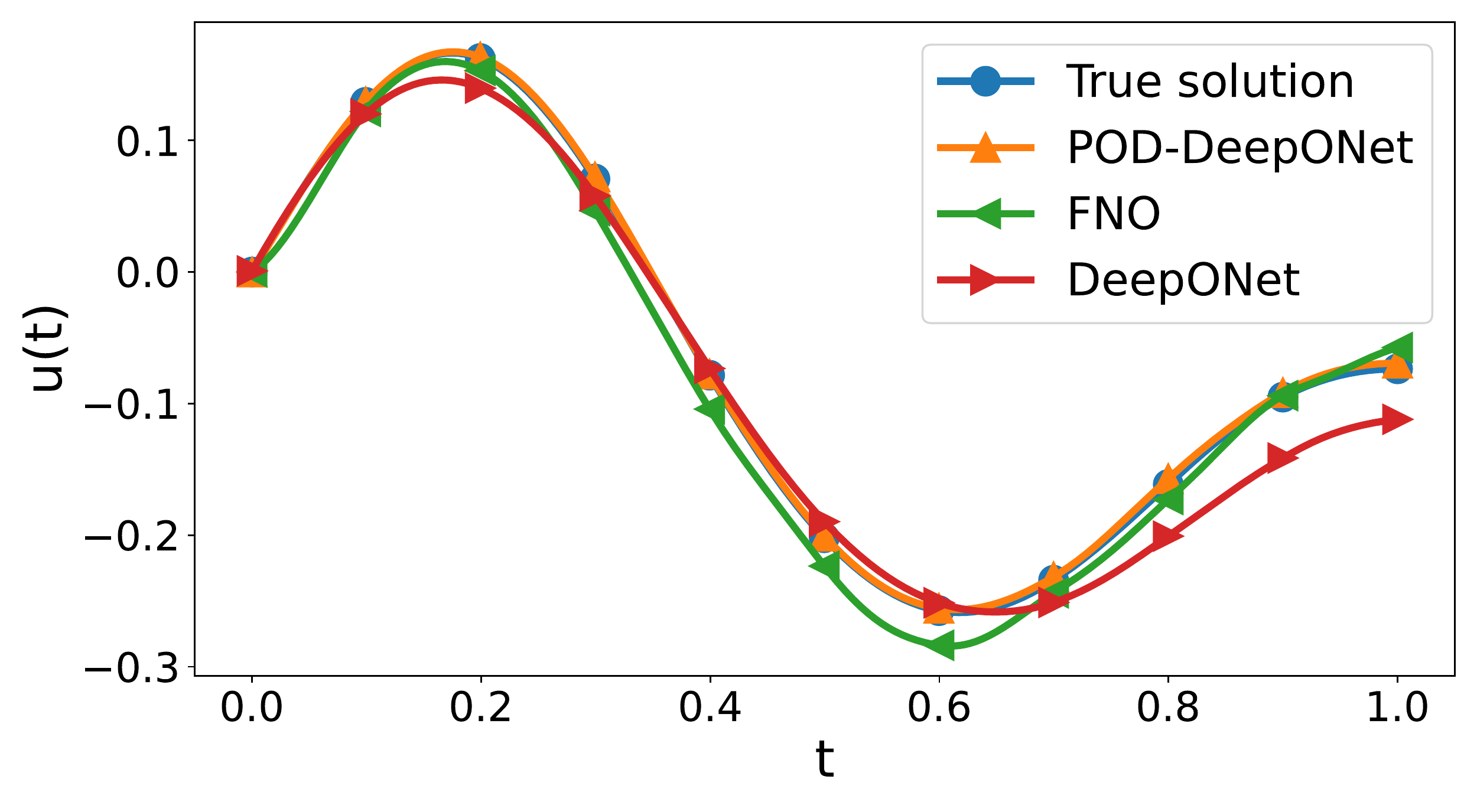}
    \end{subfigure}
    % \end{tabular}
    \caption{(Top) the forcing terms used to generate the training data for the pendulum ODE. 
    (Middle) comparing the 
    ODE solutions obtained from our model and SINDy for one of the test forcing terms. (Bottom) comparing the 
    ODE solutions obtained from POD-DeepONet, FNO, and DeepONet for one of the test forcing terms. 
    Results were obtained with training set of size 20.} 
    \label{fig:source-terms}
\end{figure}

\begin{table}[htp] 
    \centering
	\scriptsize
%	\begin{subtable}{}
		\centering
		\begin{tabular}[c]{ccccc}
%			\toprule
			\multicolumn{2}{c}{Hyper-parameters} & $I=10$  & $I=20$ & $I=20$ (Noisy)\\
			\hline \hline  
			\multirow{1}{*}{(i):}  & $\lambda_\mU$ &  $1.0e^{-8} $ &  $1.0e^{-8}$&  $1.0e^{-8} $\\ \hline
                       $u_1$            & $\sigma$ &  \multicolumn{2}{c}{$(0.15, 0.45)$} & $(0.15, 0.65)$ \\
                       $u_2 $          & $\sigma$ &  \multicolumn{2}{c}{$(0.1, 0.4)$} & $(0.1, 0.8)$ \\
            \hline \hline 
			\multirow{1}{*}{(ii):} & $\lambda_\mK$ & $1.0e^{-5} $ &  $1.0e^{-5} $&  $1.0e^{-1} $\\
            \hline
             %   nugget for $K_{f1}$ & $1.0e^{-4} $ &  $1.0e^{-5} $&  $1.0e^{-1} $\\
              \multirow{3}{*}{ $\overline{\mP}_1$ }   & $\ell_1 =\ell_2$  & $0.52 $ &  $1.0 $&  $1.0 $\\
                 & $d$   & $5 $ &  $3 $&  $1 $\\
                 & $c$  & $3.5 $ &  $0.015 $&  $0.01 $\\
                 \hline
               \multirow{3}{*}{ $\overline{\mP}_1$ }  & $\ell_1 = \ell_2$ & $3.0 $ &  $2.4 $&  $1.9 $\\
                 & $d$  & $5 $ &  $3 $&  $1 $\\
                 & $c$  & $2.8 $ &  $0.01 $&  $0.01 $\\
%			\bottomrule
		\end{tabular}
%	\end{subtable}\\
	\caption{The hyper-parameters we used for the pendulum experiments for exact training 
 sets of size $I=10, 20$ as well as the noisy  training set of size $I=20$.  
 % has two ODEs, so we used the same two kernels, the ARD kernel or the polynomial kernel to learn the two operators. All the training solutions were interpolated once, so the ARD kernel and the polynomial kernel used the same derivatives as inputs.
 } \label{tb:hyperpar-pendulum}
	\vspace{0.0in}
\end{table}

\subsection{Diffusion PDE}\label{sec:diffusion-details}
The test data set was generated by drawing the source terms from the same 
GP as in the pendulum example of Section~\ref{sec:pendulum-details}.
The solution $u^{(i)}(x,t)$ for each force $f^{(i)}(x)$ 
was computed on a fine grid using an independent finite difference solver
before they were subsampled to a space-time grid of size $15 \times 15$, constituting the training set, so for each tuple $(u^{(i)}, f^{(i)} )$ we collected a 
total of 225 values for a total training set size of  $I=10$ and $20$. The test data set was produced in the same manner 
for 50 pairs of solutions and sources. 
The errors were once again computed by averaging the $L^2$ errors over the 
test set. 
% The training data was pre-processed using our kernel smoothing approach
%  with the kernel $\mU$ 
% taken to be the RBF kernel with length scale 0.53 and nugget $\beta_\mU^2 = 10^{-10}$ chosen via CV. 
% The function $\overline{\mP}$ was learned using our approach with $\mK$ taken to be 
% the RBF kernel with length scale 20.01 and nugget $\beta_\mK^2 = 10^{-6}$ chosen via CV. 
% The SINDy algorithm was used 
% to learn the PDE with the dictionary $\{ u, u_t, u_{xx}, u^2, u^3,  u_{xx}u, u_{xx}u^2, 
% u_{xx}u^3, f, 1\}$. We used the same threshold levels as the pendulum example.
 When implementing SINDy we used the dictionary of functions  $\{ u_t, u_{xx}, u, u^2, u^3, u\cdot u_{xx}, u^2\cdot u_{xx}, u^3\cdot u_{xx}, u\cdot u_t,
 u^2\cdot u_t, u^3\cdot u_t,1 \}.$

 We parameterized the PDE as 
 \begin{equation*}
 \overline{\mP}(u(x,t), u_t, u_{xx}(x,t) ) = f(x).
 \end{equation*}
 All hyperparameters involved in the training of  our kernel method for this 
 example are summarized in Table~\ref{tb:hyperpar-diffusion}. Once again 
 we used the Gaussian kernel for Step (i) while the ARD and polynomial kernels 
 were used for Step (ii). Step (iii) also used the Gaussian kernel with a lengscale that 
 was chosen in the same range that was found in Step (i).
 We also present an example of the predicted solutions of the PDE from the 
 test set in Figure~\ref{fig:diffusion-solution}.

\begin{table}[htp] 
    \centering
	\scriptsize
%	\begin{subtable}{}
		\centering
  		\begin{tabular}[c]{ccccc}
%			\toprule
			\multicolumn{2}{c}{Hyper-parameters} & $I=10$  & $I=20$ & $I=20$ (Noisy)\\
			\hline \hline  
			\multirow{1}{*}{(i):}  & $\lambda_\mU$ & $1.0e^{-3} $ &  $1.0e^{-3} $&  $1.0e^{-3} $\\
                                & $\sigma$ &  \multicolumn{2}{c}{$(0.15, 0.7)$} & $(0.4, 1.0)$ \\
            \hline \hline 
			\multirow{1}{*}{(ii):} & $\lambda_\mK$ & $1.0e^{-3} $ &  $1.0e^{-3} $&  $1.0e^{-3} $\\
 
             %   nugget for $K_{f1}$ & $1.0e^{-4} $ &  $1.0e^{-5} $&  $1.0e^{-1} $\\
                 & $(\ell_1, \dots, \ell_3)$  
              & $(0.50,1.3,0.13 ) $ &  $(0.50,1.3,0.13 )$&  $(0.50,2.0,0.25 ) $\\
                 & $d$  & $2 $ &  $2 $&  $Failed $\\
                 & $c$  & $0.23 $ &  $0.0 $&  $Failed $\\
%			\bottomrule
		\end{tabular}
% 		\begin{tabular}[c]{cccc}
% %			\toprule
% 			Hyper-parameters & 10 sources & 20 sources & Noisy 20 sources\\
% 			\hline
% 			$\lambda_\mU$ &  \\
% 			$\lambda_\mK$ & $1.0e^{-3} $ &  $1.0e^{-3} $&  $1.0e^{-3} $\\
%                 ARD length scale for $K_f$ & $(0.50,1.3,0.13 ) $ &  $(0.50,1.3,0.13 )$&  $(0.50,2.0,0.25 ) $\\
%                 d for polynomial kernel $K_f$ & $2 $ &  $2 $&  $Failed $\\
%                 c for polynomial kernel $K_f$ & $0.23 $ &  $0.0 $&  $Failed $\\
% %			\bottomrule
% 		\end{tabular}
%	\end{subtable}\\
	\caption{The hyper-parameters we used for the diffusion experiment 
 for exact training sets of size $I=10,20$ as well as the noisy training set of size 
 $I=20$. Reported "Failed" values indicate high errors that were not competitive.} \label{tb:hyperpar-diffusion}
	\vspace{0.0in}
\end{table}

\begin{figure}[htp]
    \centering
    \includegraphics[width= 0.45 \textwidth]{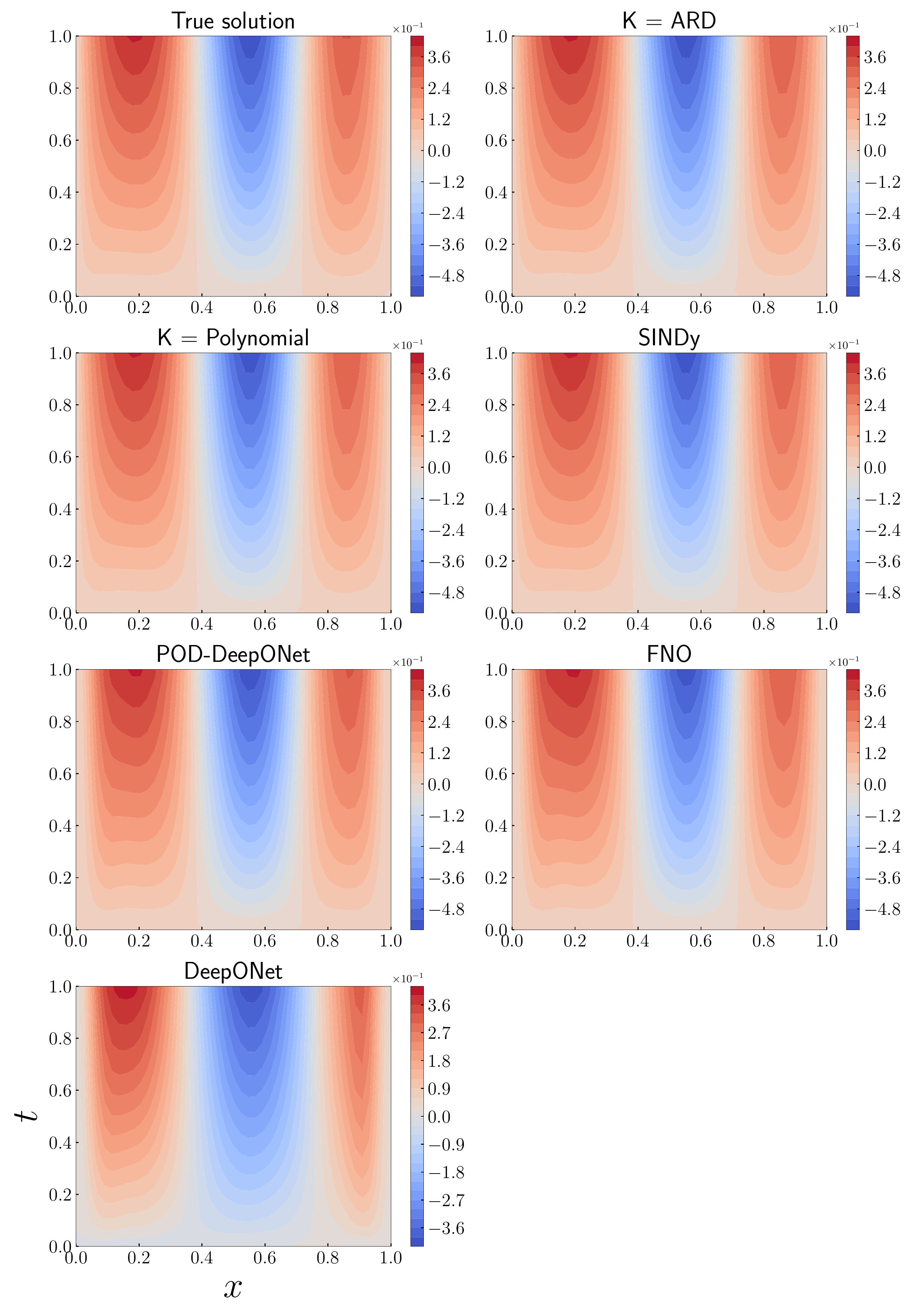}
    \caption{A comparison of the estimated solutions to the diffusion PDE for one of the 
    forcing terms in the test set with training set of size $I=20$.}
    \label{fig:diffusion-solution}
\end{figure}

\subsection{Darcy Flow}\label{seC:darcy-details}
The training and test sources for the Darcy flow PDE were generated 
by taking $f(x_1, x_2) \equiv f(x_2)$ and  drawing this function 
from a 1D GP with the RBF kernel and  length scale $0.2$. The PDE 
was then solved using a finite difference solver, on a fine mesh and the solutions 
were  subsampled to a uniform grid of size $15 \times 15$, following a 
similar scheme to the diffusion PDE. The test set was generated in the same manner. 

We parameterized the PDE as 
\begin{equation*}
    \overline{\mP}(x_1, x_2, u, u_{x_1}, u_{x_2}, \Delta u ) = f(\bx).
\end{equation*}
All hyperparameters involved in the training of our kernel method for this example as summarized in Table~\ref{tb:hyperpar-Darcy}. The Gaussian kernel was used for Step (i)
while the ARD kernel was used for Step (ii). Our experiments using the 
polynomial kernel for this step lead to bad results. Step (iii) also used the Gaussian kernel
with a lengthscale that was chosen in the range that was tuned in Step (i). Example solutions from the test set are presented in Figure~\ref{fig:darcy-solution}.

\begin{figure}
    \centering
    \includegraphics[width=0.44 \textwidth]{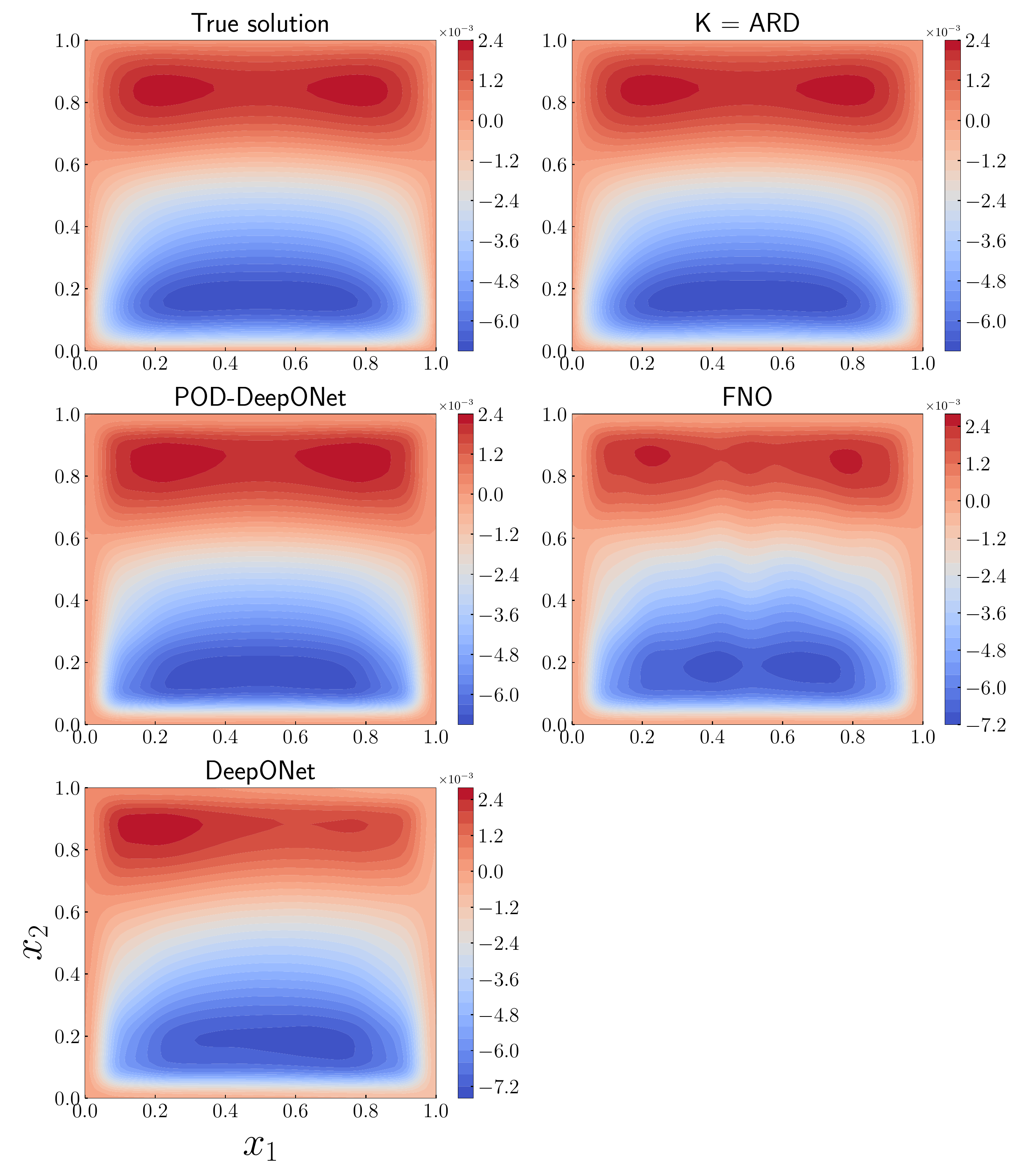}
    \caption{A comparison of the estimated solutions to the Darcy flow PDE for one of the 
    forcing terms in the test set (the 20 training source experiment).}
    \label{fig:darcy-solution}
\end{figure}

\begin{table*}[htp] 
    \centering
	\scriptsize
%	\begin{subtable}{}
		\centering
    		\begin{tabular}[c]{ccccc}
%			\toprule
			\multicolumn{2}{c}{Hyper-parameters} & $I=10$  & $I=20$ & $I=20$ (Noisy)\\
			\hline \hline  
			\multirow{1}{*}{(i):}  & $\lambda_\mU$ &  $1.0e^{-8} $ &  $1.0e^{-8} $&  $1.0e^{-2} $\\
                                & $\sigma$ &  \multicolumn{2}{c}{$(0.15,0.35)$} & $(0.05, 0.5)$ \\
            \hline \hline 
			\multirow{1}{*}{(ii):} & $\lambda_\mK$ & $1.0e^{-3} $ &  $1.0e^{-3} $&  $1.0e^{-1} $\\
             %   nugget for $K_{f1}$ & $1.0e^{-4} $ &  $1.0e^{-5} $&  $1.0e^{-1} $\\
                 & $(\ell_1, \dots, \ell_6)$  
               & $(1.2,1.2,8.0,8.0,10.0,10.0) $ &  $(0.4,0.4,3.2,3.2,5.0,5.0)$&  $(0.64,0.64,2.0,2.0,3.0,3.0) $\\
%			\bottomrule
		\end{tabular}
% 		\begin{tabular}[c]{cccc}
% %			\toprule
% 			Hyper-parameters & 10 sources & 20 sources & Noisy 20 sources\\
% 			\hline
% 			$\lambda_\mU$ &  $1.0e^{-8} $ &  $1.0e^{-8} $&  $1.0e^{-2} $\\
% 			 $\lambda_\mK$ & $1.0e^{-3} $ &  $1.0e^{-3} $&  $1.0e^{-1} $\\
%                 $\ell_1,\dots, \ell_6$ & $(1.2,1.2,8.0,8.0,10.0,10.0) $ &  $(0.4,0.4,3.2,3.2,5.0,5.0)$&  $(0.64,0.64,2.0,2.0,3.0,3.0) $\\
%                 % d for polynomial kernel $K_f$ & $Failed $ &  $Failed $&  $Failed $\\
%                 % c for polynomial kernel $K_f$ & $Failed $ &  $Failed $&  $Failed $\\
% %			\bottomrule
% 		\end{tabular}
%	\end{subtable}\\
	\caption{The hyperparameters we used for the Darcy Flow experiments for 
 exact training sets of size $I=10, 20$ as well as noisy 
 training set of size $I=20$.} \label{tb:hyperpar-Darcy}
	\vspace{0.0in}
\end{table*}

% \begin{thebibliography}{}
% \setlength{\itemindent}{-\leftmargin}
% \makeatletter\renewcommand{\@biblabel}[1]{}\makeatother
% \bibitem{} J.~Alspector, B.~Gupta, and R.~B.~Allen (1989).
%     \newblock Performance of a stochastic learning microchip.
%     \newblock In D. S. Touretzky (ed.),
%     \textit{Advances in Neural Information Processing Systems 1}, 748--760.
%     San Mateo, Calif.: Morgan Kaufmann.

% \bibitem{} F.~Rosenblatt (1962).
%     \newblock \textit{Principles of Neurodynamics.}
%     \newblock Washington, D.C.: Spartan Books.

% \bibitem{} G.~Tesauro (1989).
%     \newblock Neurogammon wins computer Olympiad.
%     \newblock \textit{Neural Computation} \textbf{1}(3):321--323.
% \end{thebibliography}

\end{document}